\documentclass[ijds,sglanonrev]{informs4}
\RequirePackage{tgtermes}
\RequirePackage{newtxtext}
\RequirePackage{newtxmath}
\RequirePackage{bm}
\RequirePackage{endnotes}
\usepackage{url}

\OneAndAHalfSpacedXII 

\usepackage{algorithm}
\usepackage{algpseudocode}
\usepackage{tikz}
\usepackage{xcolor}         
\usepackage{tikz}
\usetikzlibrary{positioning,shapes.geometric,fit,backgrounds}
\usepackage{booktabs}
\usepackage{amsmath}
\allowdisplaybreaks
\usepackage{hyperref}
\usepackage{subcaption}
\usepackage{wrapfig}
\usepackage{listings}
\usepackage{clipboard}
\usepackage[most]{tcolorbox}

\usepackage{natbib}
 \bibpunct[, ]{(}{)}{,}{a}{}{,}%
 \def\bibfont{\small}%

\usepackage{tabularx}
\usepackage[margin=1in]{geometry}
\usepackage{pgfplots}
\usepackage{pgfplotstable}
\usepgfplotslibrary{groupplots,fillbetween}
\pgfplotsset{compat=1.18}
\EquationsNumberedThrough    

\TheoremsNumberedThrough     
\ECRepeatTheorems  %

\MANUSCRIPTNO{IJDS-0001-2024.00}

\newclipboard{response-snippets}
\newtcolorbox{revisionquote}{
  enhanced,
  breakable,
  sharp corners,
  boxrule=0.4pt,
  colback=white,
  colframe=blue,
  coltext=blue, 
  borderline west={2pt}{0pt}{blue}
}

\begin{document}

\RUNAUTHOR{Zhang et al.}

\RUNTITLE{Comparing E\&E Strategies of LLMs and Humans}

\TITLE{Comparing Exploration–Exploitation Strategies of LLMs and Humans: Insights from Standard Multi-armed Bandit Experiments}

\ARTICLEAUTHORS{%
\AUTHOR{Ziyuan Zhang}
\AFF{Department of Mechanical \& Industrial Engineering,
University of Toronto, \EMAIL{zyuan.zhang@mail.utoronto.ca}}

\AUTHOR{Darcy Wang}
\AFF{Department of Mechanical \& Industrial Engineering,
University of Toronto, \EMAIL{darcy.wang@mail.utoronto.ca}}

\AUTHOR{Ningyuan Chen}
\AFF{The Rotman School of Management,
University of Toronto, \EMAIL{ningyuan.chen@utoronto.ca}}

\AUTHOR{Rodrigo Mansur}
\AFF{Department of Psychiatry,
University of Toronto, \EMAIL{Rodrigo.Mansur@uhn.ca}}

\AUTHOR{Vahid Sarhangian}
\AFF{Department of Mechanical \& Industrial Engineering,
University of Toronto, \EMAIL{sarhangian@mie.utoronto.ca}}
} 

\ABSTRACT{%
 Large language models (LLMs) are increasingly used to simulate or automate human behavior in complex sequential decision-making settings. A natural question is then whether LLMs exhibit similar decision-making behavior to humans, and can achieve comparable (or superior) performance. 
In this work, we focus on the exploration-exploitation (E\&E) tradeoff, a fundamental aspect of dynamic decision-making under uncertainty. We employ canonical multi-armed bandit (MAB) experiments introduced in the cognitive science and psychiatry literature to conduct a comparative study of the E\&E strategies of LLMs, humans, and MAB algorithms. We use interpretable choice models to capture the E\&E strategies of the agents and investigate how enabling thinking traces, through both prompting strategies and thinking models, shapes LLM decision-making. We find that enabling thinking in LLMs shifts their behavior toward more human-like behavior, characterized by a mix of \emph{random} and \emph{directed} exploration. In a simple stationary setting, thinking-enabled LLMs exhibit similar levels of random and directed exploration compared to humans. However, in more complex, non-stationary environments, LLMs struggle to match human adaptability, particularly in effective directed exploration, despite achieving similar regret in certain scenarios.
    Our findings highlight both the promise and limits of LLMs as simulators of human behavior and tools for automated decision-making and point to potential areas for improvement. 
}%





\maketitle

\section{Introduction}\label{sec:Intro}
Large Language Models (LLMs), while originally developed for text generation, are increasingly used as simulators of human behavior. Researchers across disciplines have leveraged LLMs to simulate how humans respond to different stimuli, such as consumer choices and preference \citep{aher2023using, arora2025ai, goli2024frontiers, brand2023using}, political messaging \citep{argyle2023out}, moral context \citep{dillion2023can}, and even interactive daily social routines \citep{park2023generative}. Additionally, there is growing interest in using LLMs to support or automate decision-making in complex sequential settings, e.g., \citep{hao2023reasoning, liu2024reason, huang2024wese}. 
Recent advances in explicit thinking techniques, which generate a deliberation trace before producing a final response, such as Chain-of-Thought prompting \citep{wei2022chain, kojima2022large}, reinforcement learning, and tool-augmented reasoning (see \citep{xu2025towards} and \citep{ke2025survey} for recent surveys) have substantially improved LLM performance on multi-step reasoning tasks (e.g., mathematical problem solving, code generation), thus potentially expanding their ability to simulate or automate dynamic decision-making tasks in real-world settings. However, it remains unclear whether LLMs can sufficiently capture the dynamic, adaptive, and cognitively grounded aspects of human decision-making, or achieve human-level (or better) performance. This motivates the central research question of our work:

    \emph{Can LLMs reliably simulate human behavior and support automated decision-making in sequential tasks under uncertainty?}

In this paper, we focus on a fundamental aspect of decision-making, namely the \emph{exploration-exploitation} (E\&E) tradeoff, arising whenever individuals must choose between leveraging known rewarding options (exploitation) and seeking out new possibilities that might lead to better outcomes (exploration). In cognitive science and psychiatry, E\&E behavior has been extensively studied to understand normative and impaired decision processes \citep{cohen2007should, daw2006cortical}, highlighting their central role in driving human behavior. A common experimental paradigm for studying this tradeoff is the (computerized) \emph{multi-armed bandit (MAB) experiments}, in which participants repeatedly choose among several options with unknown reward distributions. 
Whereas the extensive literature in computer science, economics, operations research, and statistics focuses on the design and regret analysis for MAB algorithms (see, e.g., \citealt{bubeck2012regret,lattimore2020bandit}), the cognitive science literature employs MAB experiments to understand how humans navigate the E\&E tradeoff. 
A key finding in this literature is that humans use a mixture of random and directed exploration. 
In \emph{random exploration}, individuals sample options unpredictably, introducing stochasticity into choice behavior without explicit consideration of uncertainty. In contrast, \emph{directed exploration} involves systematically selecting options associated with higher uncertainty in order to maximize information gain \citep{wilson2014humans, wilson2021balancing}. These findings underscore that human exploration is not purely random, but often goal-directed and informed by subjective uncertainty estimates and internal models of the environment. Systematic deviations from this pattern have been observed in various psychiatric populations, such as reduced directed exploration in depression \citep{smith2022lower} and increased random exploration in attention-deficit/hyperactivity disorder (ADHD) \citep{addicott2021attention} or schizophrenia \citep{cathomas2021increased}.

We investigate how LLMs navigate the E\&E tradeoff by adopting two widely-used MAB experiments from the cognitive science literature \citep{daw2006cortical, Samuel2019data}. We use standard interpretable choice models to characterize the underlying decision strategies of three commercial LLM series and compare them with those of human agents as well as MAB algorithms. We use the same experiments and with matched conditions across the agents to facilitate comparison. Previous studies (e.g.,
\citep{binz2023using, harris2025should, krishnamurthy2024can}) have examined E\&E trade-offs in LLMs, but using simple 2-armed setting. A key aspect of our work is investigating how enabling thinking in LLMs, achieved either via explicit prompting \citep{yao2023tree, wei2022chain, kojima2022large, zhang2022automatic} or thinking models that produce a thinking trace before the final response (e.g., \citep{guo2025deepseek, comanici2025gemini}), impacts the decision-making behavior of LLMs. Specifically, we carefully design two conditions to isolate the thinking capabilities. For basic models, we control for both the prompt format (with or without chain-of-thought instructions). For thinking models, we hold the prompt constant and compare with the built-in thinking module toggled on versus off. Our main contributions and findings can be summarized as follows.

First, we conduct a comprehensive comparative study of the E\&E strategies of commercial LLMs, human participants, and MAB algorithms using two standard MAB experiments: a stationary 2-armed experiment, and a more complex 4-armed non-stationary experiment. For LLM agents, we focus on \textsc{ChatGPT}, \textsc{Gemini}, \textsc{DeepSeek}, three of the most widely deployed state-of-the-art models, ensuring both practical relevance and broad applicability of our findings. Although the behavior of LLMs and algorithmic baselines can be simulated, as done in earlier studies, a key contribution of our work is the use of two datasets collected from human participants performing the same MAB tasks under controlled experimental conditions. This enables more reliable and fair comparisons between human and artificial agents. 

Second, we estimate and compare the behavior of the agents in terms of directed and random exploration, among other parameters. This allows us to investigate whether LLMs use a mixed exploration strategy as in humans and how thinking capabilities impact E\&E strategies. We find that LLMs exhibit different E\&E strategies from humans and MAB algorithms, in terms of both the overall level and the types of exploration. Thinking capabilities increase the direct exploration of LLMs in both experiments, and allow them to reach human levels for the stationary experiment. However, in the more complex non-stationary experiment, LLMs fail to engage in effective directed exploration compared to humans. 
    
Third, we compare the agents in terms of model-free performance metrics of regret and exploitation rate over time.
    This allows us to investigate how the observed differences in the E\&E strategies translate to performance gaps. We find that basic LLMs can achieve similar regret compared to humans in the stationary experiments. However, they exhibit more variable exploitation rates in the non-stationary setting and accumulate regret much faster than humans. Thinking capabilities significantly enhance the performance of LLMs, bringing their regrets to levels comparable to those of humans in certain scenarios. 
    
\section{Related Work}\label{sec:lit_review}

\paragraph{\textbf{E\&E tradeoffs in human decision-making.}} Investigating the E\&E strategies of humans and how various interventions or disorders impact them has been an active area of research in the psychiatry literature \citep{cathomas2021increased}, as well as in cognitive psychology \citep{schulz2020finding} and neuroscience  \citep{cohen2007should}. MAB experiments are a widely used experimental paradigm for studying the E\&E behavior in humans. Variants include non-stationary 4-armed bandit setting with decaying reward (\citep{daw2006cortical, chakroun2020dopaminergic, jepma2010role, beharelle2015transcranial, bornstein2017reminders, cathomas2021increased}), stationary 2-armed bandit setting \citep{Samuel2019data}, and horizon tasks comparing exploration under different information conditions \citep{wilson2014humans, wilson2021balancing}. Interpretable choice models that integrate learning and choice rules are central to characterizing human E\&E strategies. Learning rules, such as a Bayesian learner model \citep{daw2006cortical} or exponential smoothing \citep{bornstein2017reminders}, estimate the expected outcomes for each option as the experiments evolve. Choice rules then map these estimates into decisions, typically employing softmax functions \citep{daw2006cortical, chakroun2020dopaminergic} or probit regression \citep{Samuel2019data}. 
In this work, we use these models to investigate the E\&E strategies of LLMs and compare them with humans in the same tasks. 

Recent work by \cite{ding2025behavioral} models human exploration in $N$-armed bandits by defining \emph{over-} and \emph{under} exploration relative to a regret-optimal algorithm (Thompson Sampling). Rather than evaluating exploration behavior against an algorithmic notion of optimality, we aim to systematically compare humans and LLMs in how they carry out exploration. 
This distinction becomes particularly important in non-stationary environments, where the regret-optimality guarantees of classical algorithms no longer apply. 

Similarly, \cite{zhuo2025navigating} applies structural estimation to observational data to identify algorithmic policies (e.g., UCB rules) that explain human decision-making. In contrast, our experimental design features directly observed rewards, isolating exploration behavior without requiring latent preference inference. Furthermore, instead of finding the best-fitting algorithmic policy, we leverage a behavioral choice model to directly compare the underlying E\&E mechanisms of humans and LLMs.

\paragraph{\textbf{Understanding the decision-making behavior of LLMs.}} 
Several studies have investigated various aspects of decision-making behavior for LLMs, including risk preferences and loss aversion \citep{jia2024decision, binz2023using}, and trust behavior \cite{xie2024can}. \cite{binz2023using} studies several behavioral aspects of \textsc{GPT3} using standard tasks from cognitive psychology, including the Horizon task (a variant of the 2-armed MAB experiment) to investigate E\&E strategies. Our study expands that of \cite{binz2023using} in several ways using the more complex 4-armed setting with decaying rewards, and considering other agents. Consistent with their findings, we also find that LLMs engage in limited directed exploration and that LLMs can have low-regret in the stationary 2-armed setting. In contrast, we find that LLMs can have large regret in the more complex 4-armed setting, but thinking capabilities can significantly boost their exploration capabilities and lower regret. A related body of work focuses on understanding the role-behavior capabilities of LLMs 
(i.e., impersonating different human roles). \cite{salewski2023context, shen2024decision} examine the decision-making behavior of LLMs in-context, including their E\&E strategies using 2-armed MAB experiments. 

\paragraph{\textbf{LLMs for optimal dynamic decision-making.}}
A growing body of literature investigates the performance of LLMs in dynamic or sequential decision-making problems (e.g., \citep{hao2023reasoning, liu2024reason, huang2022inner, raparthy2023generalization, li2022pre, huang2024wese, park2024llm}). Because managing the E\&E tradeoff is a key component of optimal dynamic decision-making, several studies have explored and aimed to improve LLMs’ capabilities in MAB problems, in particular focusing on their ability to make optimal decisions through explorations \citep{krishnamurthy2024can, harris2025should, nie2024evolve}. 
In contrast, we focus on understanding the E\&E strategies of LLMs in MAB experiments and, in particular, how they differ from those of human agents. 
We also show that thinking capabilities could substantially improve the performance of LLMs in terms of regret.

\paragraph{\textbf{LLMs for simulating human behavior.}}There is a growing interest in leveraging LLMs to simulate human behavior, also known as the digital twins simulation, across diverse disciplines, including cognitive science \citep{dillion2023can}, computer science \citep{aher2023using, park2024generative}, and political science \citep{argyle2023out}. In the social sciences, such as economics and marketing, researchers have also examined the use of LLMs to model consumer decision-making and agent behavior in strategic or preference-based settings (e.g., \citealt{horton2023large, arora2025ai, brand2023using, wang2024large}). Recent work on agentic e-commerce investigates how AI agents autonomously browse, evaluate, and transact in online marketplaces \citep{allouah2025your}. Recent large-scale data collection efforts, such as \citet{toubia2025twin}, provide rich individual-level benchmarks for constructing digital twins of thousands of participants, further expanding the empirical foundation for this line of research. While these efforts are promising, concerns remain regarding the reliability and validity of employing LLMs as human simulators, with recent work highlighting key challenges and limitations \citep{goli2024frontiers, gui2023challenge, huang2024social}. Our comparative analysis offers insights into whether LLMs exhibit similar exploration–exploitation strategies compared to humans, a central dimension of decision-making across many of the above domains.

To better position our work, we provide Table \ref{tab:positioning} that summarizes the key differences of our work and previous literature that examines the E\&E behavior of LLMs in bandit tasks.
\begin{table}[ht]
\centering
\small
\caption{Literature examines E\&E behavior of LLMs}
\label{tab:positioning}
\begin{tabularx}{\textwidth}{l X c X} 
\toprule
\textbf{Paper} & \textbf{Behavioral Task(s)} & \textbf{Human Data} & \textbf{Analysis Approach}\\
\midrule
\cite{binz2023using} & 6-rounds Horizon task & Y & Cognitive psychology experiments \\ \addlinespace
\cite{krishnamurthy2024can}  & Stochastic Bernoulli bandits & N & Surrogate statistics \\ \addlinespace
\cite{nie2024evolve} & Bernoulli/Gaussian (contextual) bandits & N & Fine-tuning on traces from optimal algorithms \\ \addlinespace
\textbf{This paper} & Stationary 2-armed \& nonstationary 4-armed bandits & Y & Behavioral/structural  models \\
\bottomrule
\end{tabularx}
\end{table}
To summarize, unlike prior studies that focus on evaluating or optimizing LLMs for algorithmic optimality, we focus on comparison with human decision making in sequential tasks. To the best of our knowledge, this is the \emph{first study} to provide a direct comparison between LLMs and human behavior under identical and complex bandit settings, including in non-stationary environments. While \cite{binz2023using} also incorporate human data, their analysis of exploration behaviors relies on traditional cognitive psychology experiments limited to the Horizon task. In contrast, our approach leverages behavioral and structural models to systematically investigate decision-making processes across both stationary 2-armed and non-stationary 4-armed bandit settings.

Although similar to the prior literature we find that LLMs may fail to exhibit efficient exploration, we also show that these limitations can be addressed by their reasoning/thinking abilities in stationary bandit settings without fine-tuning. In contrast, in non-stationary environments, human agents demonstrate greater adaptability than LLMs as well as standard algorithmic baselines and achieve the lowest regret. 

Overall, we utilize a behavioral modeling framework that enables a principled and interpretable comparison of human and LLM decision-making processes. This framework provides a systematic understanding of their differences and offers insights on designing LLMs that behave more adaptively and more closely resemble human sequential decision-making.

\section{Experimental Setup}\label{sec:setup}

In this section, we describe the experimental setup used to evaluate and compare agents' behavior. We detail the design of the two bandit experiments, the prompt design procedures for eliciting LLM decisions, and the human datasets and algorithmic benchmarks used in the experiments. The code for reproducing all experiments is available at \url{https://github.com/zzy620/LLM-exploration-exploitation}.

\subsection{Multi-armed Bandit (MAB) Experiments}\label{sec:mab}
We begin by formally defining a general MAB experiment. There are $K \in \mathbb{N}$ available arms (actions). The decision-maker interacts with the environment over $T$ discrete time steps, or \emph{\textbf{rounds}}, indexed by $t\in [T]:=\{1,\ldots,T\}$. Each arm $k \in [K]$ has a possibly non-stationary reward distribution $P_k(t)$ that is independent of other arms. At each round $t$, the decision-maker selects an arm $a_t \in [K]$ according to some policy $\pi$ and receives a reward $r_{t}\sim P_{a_t}(t)$. This generates a trajectory of actions and rewards $\{a_1,r_1,\dots, a_T, r_T\}$ for a single $\emph{trial}$ of the experiment. We consider two variants of MAB experiments from the literature, which we briefly introduce next.

\paragraph{\textbf{Stationary 2-armed bandit.}} The first experiment is originally proposed in \cite{Samuel2019data} and subsequently used in several studies, e.g., \cite{shen2024decision, wilson2021balancing, salewski2023context}.
In this setting, each agent completes 20 trials of a 2-armed bandit experiment with $T = 10$ rounds in each trial. 
Both arms generate stochastic (but stationary) rewards. 
At the start of each trial, the mean rewards for two arms, $\mu_1$ and $\mu_2$, are independently sampled from a Gaussian distribution $\mathcal{N}(0, 100)$. 
In each round, the reward associated with arm $k\in\{ 1,2\}$ is drawn from $\mathcal{N}(\mu_k, \sigma^2_0)$, with $ \sigma_0^2 \equiv 10$ fixed across arms and rounds. 

\paragraph{\textbf{Non-stationary 4-armed bandit.}} The second experiment is originally proposed in \citep{daw2006cortical} and has been widely adopted in the cognitive psychiatry literature, e.g., \citep{chakroun2020dopaminergic, jepma2010role, beharelle2015transcranial, bornstein2017reminders}. To encourage agents' exploration behavior, this experiment introduces non-stationary mean rewards that follow a diffusion process. 
Each subject completes one trial of the experiment with $T = 300$ rounds \citep{chakroun2020dopaminergic}, or two trials with $T = 150$ rounds \citep{daw2006cortical}. 
The reward in round $t$ of arm $k$ is drawn from a Gaussian distribution with a time-varying mean $\mu_{k,t}$, and a fixed variance $\sigma_0^2$, i.e.,
    $r_{k,t} \sim \mathcal{N}(\mu_{k,t}, \sigma_0^2)$.
The variance of the reward is set to $\sigma_0^2 = 4$. 
The mean reward for each arm follows a decaying Gaussian random walk given by \begin{equation}
    \mu_{k,t+1} = \lambda \mu_{k,t} + (1-\lambda)\theta + \omega_t,
\end{equation}
for $t\in\{1,\ldots,T-1\}$ with $\mu_{1,1} = 20, \mu_{2,1} = 40, \mu_{3,1} = 60, \mu_{4,1} = 80$. 
In the formula, $\lambda$ and $\theta$ represent the decaying rate and long-term mean, respectively and are set to $\lambda = 0.9836$ and $\theta = 50$. 
Furthermore, $\omega_t$ denotes the diffusion noise and is drawn independently in each round from a zero-mean Gaussian distribution with variance $\sigma_d^2=2.8$. Different from the 2-armed bandit, three groups of reward sequences (for all arms) are \emph{pre-generated} before the experiment. 
Each participant is then randomly assigned to one of the reward groups.

\subsection{Agents Completing the MAB Tasks}\label{sec:agents}
In this section, we introduce the agents that have completed the experiments, whose data have been recorded.
They include three categories: human participants, LLM agents, and popular MAB algorithms. 

\paragraph{\textbf{Human participants.}}
We use two datasets from previous studies collected from human participants.
The first dataset is sourced from \cite{Samuel2019data} and consists of 45 human participants completing the stationary 2-armed bandit experiment. 
It is collected via the Amazon Mechanical Turk (MTurk) platform\footnote{Publicly available at: \url{https://github.com/sjgershm/exploration}}. 
For each trial of each participant, the dataset includes the fixed mean rewards for both arms; for each round within a trial, it records the participant’s selected arm and the corresponding integer reward observed by the participant.
For the non-stationary 4-armed bandit task ($T=300$), we use the dataset collected from a controlled lab experiment in \cite{elifedata}\footnote{Dataset available at Zenodo upon request: \url{https://doi.org/10.5281/zenodo.3872973}}, reported in \citep{chakroun2020dopaminergic}. 
The dataset consists of 31 participants, and each participant is randomly assigned to one of three pre-generated reward groups, with all reward sequences fixed in advance. For each round, the dataset records the participant’s selected arm and the corresponding reward according to the assigned reward group.

In both MAB experiments, the participants are explicitly informed that each arm yields an uncertain reward, and their main objective is to maximize their total reward. Compensation consisted of a fixed payment for participation, with an additional performance-based bonus in the 4-armed bandit task. See Appendix~\ref{appendix:human} for more details.

\paragraph{\textbf{LLM agents.}} We employ LLMs as participants of the two experiments described above. 
 We closely follow the original experimental protocols for both settings, ensuring that LLMs receive equivalent information throughout the experiment as the human participants. We test a set of popular commercial LLMs, namely \textsc{GPT-4o-mini}, \textsc{Gemini-2.0-flash}, and \textsc{DeepSeek-V3}, along with their corresponding thinking models, namely \textsc{GPT-o3-mini}, \textsc{Gemini-2.5-flash}, and \textsc{DeepSeek-R1}. And our primary goal using these models is that these are the most widely used and state-of-the-art thinking-capable models, making the behavioral comparison with humans both relevant and practically valuable.

For LLMs to complete both MAB experiments, 
we first provide them with a system prompt describing the experiment and outlining the main objective. Similar to human participants, LLMs are instructed to aim at maximizing their total reward. 
Then, in each round, we send a prompt first summarizing the chosen arms and received rewards of past rounds in JSON format. We then instruct the LLMs to make their decisions. 

For basic (non-thinking) models (\textsc{ChatGPT-4o-mini}, \textsc{Gemini-2.0-flash}, and \textsc{DeepSeek-V3}), we test two variations of prompts: basic and Chain-of-Thought (CoT).
In the basic version, we ask them to directly make their choices without any explanation.
In the CoT version, we encourage them to \textit{``think out loud''} before making a choice. Note that our CoT prompting is slightly different from formalized prompts such as \textit{``Let’s think step by step.''} in the literature \citep{wei2022chain, kojima2022large}.
This is because the MAB experiments do not involve well-defined and step-by-step mathematical reasoning. 
Nevertheless, this approach significantly elicits interpretable thinking traces from LLMs (see Appendix \ref{appendix:prompt} for several illustrative examples).
For thinking models \textsc{ChatGPT-o3-mini} and \textsc{Gemini-2.5-flash}, we do not rely on prompting variations but instead control the model's internal thinking behavior. Specifically, we toggle the thinking mode on or off, or vary the thinking budget allocated. This allows us to systematically control and analyze the model's internal thinking trace during decision-making. Notice that although \textsc{DeepSeek-R1} is also a thinking model, it does not provide an interface to manually toggle the thinking mode or adjust the thinking budget. Therefore, we directly use \textsc{DeepSeek-R1} in our experiments without varying its internal thinking configuration.

\begin{figure}[ht]
    \centering
    \includegraphics[width=0.80\linewidth]{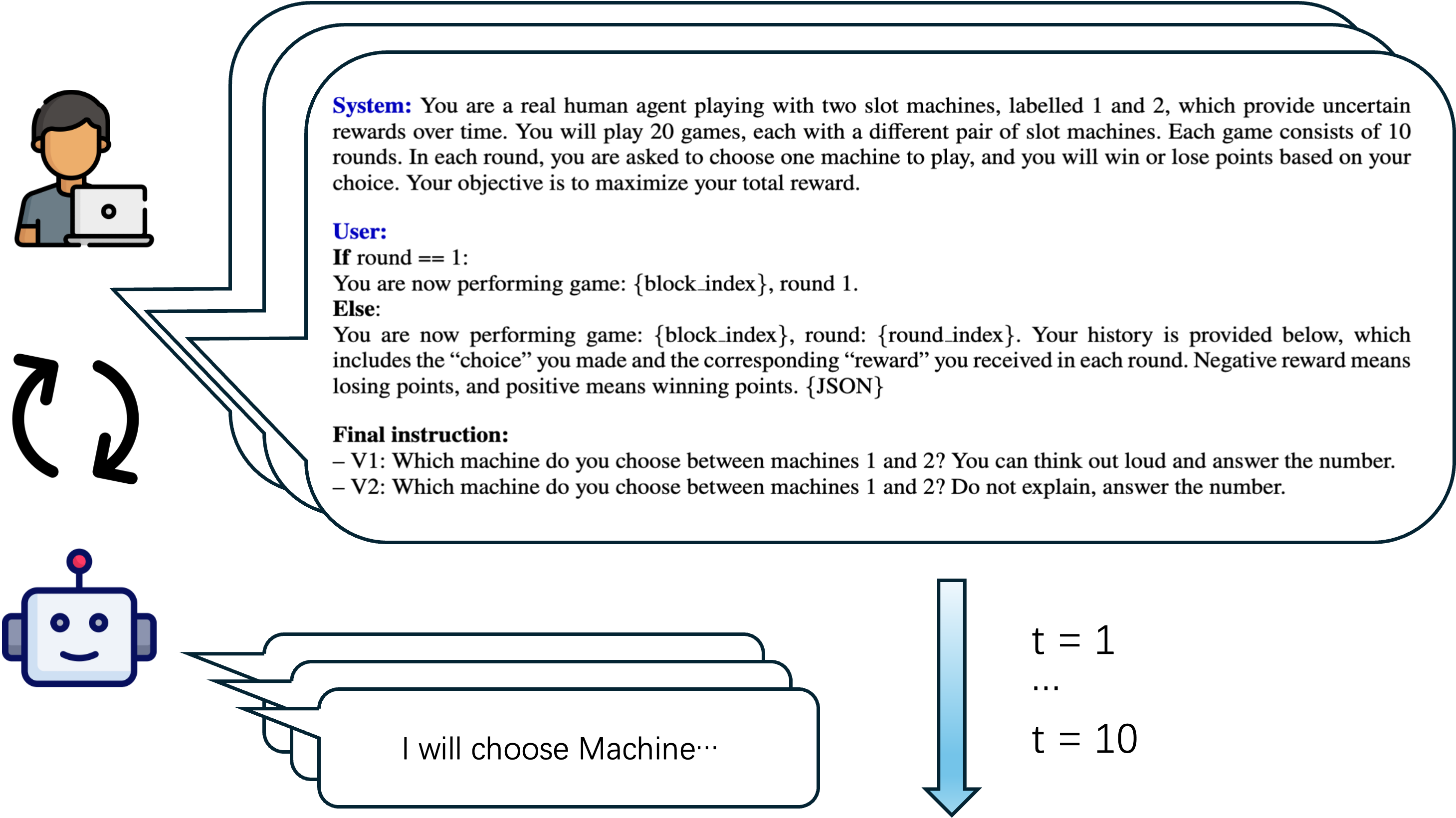}
    \caption{Illustration of the prompt protocol and interaction flow in the 2-armed bandit experiment. }
    \label{fig:placeholder}
\end{figure}

Details of prompt designs are presented in Appendix~\ref{appendix:prompt}. For each LLM model and prompting/thinking variation, we conduct 300 independent trials for the 2-armed bandit experiment and 15 for the 4-armed bandit experiment, to match the experiments for human agents.

We acknowledge that the interaction interface of the experiment differs between human subjects (visual GUI with colored button indicating each arm) and LLM agent (JSON-structured text). Text is the native modality for LLMs. Forcing them to process screenshots would introduce uncontrolled visual-perceptual errors rather than testing their strategic reasoning. By using natural language and JSON, we ensure the LLM receives the exact same information as humans, just in the format it processes most accurately. Furthermore, utilizing structured text summaries to represent information is a standard and widely accepted practice in the literature for evaluating LLM agents (e.g., \cite{krishnamurthy2024can, park2024llm}, and \cite{binz2023using})

\paragraph{\textbf{MAB algorithms.}}
We include three popular MAB algorithms, the Upper Confidence Bound (UCB) algorithm \citep{lai1985asymptotically,auer2002finite}, Thompson Sampling (TS) \citep{thompson1933likelihood}, and $\epsilon$-greedy \citep{sutton1998reinforcement}. 
See Appendix~\ref{appendix:algorithm} for the details in implementing these algorithms in our experiments.
We refer to textbooks and surveys \citep{bubeck2012regret,lattimore2020bandit} for a theoretical treatment of the algorithms such as their regret analyses and \cite{vermorel2005multi} for an empirical investigation.
In particular, all three algorithms have been shown to learn the optimal arm and achieve near-optimal regret in stationary settings with proper hyperparameters, as the time horizon increases.

\section{Choice Models and Estimation}\label{sec:choice_models}
In this section, we present the choice models used to characterize human and LLM decision-making. 
We describe two model classes based on softmax functions and probit regression, commonly used in the literature.
We do not propose new models that may fit the data better, in order to compare to and benchmark against the observations made in prior studies. We then employ a hierarchical Bayesian approach for parameter estimation.

\subsection{Choice Models}
For both model classes, we assume that agents form and update estimates of the expected value $Q_k(t)$ and standard deviation $S_k(t)$ of the mean payoff of each arm $k \in [K]$ and in each round $t$.
These quantities are computed using an established Bayesian learner model \citep{daw2006cortical} which is based on the Kalman filter \citep{anderson2005optimal, kalman1960new}. 
Suppose the subject's belief over the true mean payoff $\mu_{k,t}$ for each arm $k \in [K]$ at the beginning of round $t$ follows a prior distribution $N(Q_k(t), S^2_k(t))$. If arm $a_t$ is chosen with an observed reward $r_t$, then the posterior mean $Q_{a_t}^{pos}(t)$ and variance $S_{a_t}^{2 pos}(t)$ are updated as,
\begin{equation}
    Q_{a_t}^{pos}(t) = Q_{a_t}(t) + \kappa(t)(r_t - Q_{a_t}(t)),
\end{equation}
\begin{equation}
    S_{a_t}^{2pos}(t) = (1-\kappa(t))S_{a_t}^2(t),
\end{equation}
where $\kappa(t)$ is the Kalman gain for each round $t$ computed as
\begin{equation}
    \kappa(t) = S^2_{a_t}(t)/(S^2_{a_t}(t) + \hat{\sigma}_0^2).
\end{equation}
We set $Q_k(1) = Q(1), S_k(1) = S(1)$ for each arm $k\in[K]$ at the beginning of each trial, implying that the subjects have the same estimates over different arms. Further, for arms that are not chosen in a round, posteriors are equal to the prior in that round, i.e., $Q_k^{pos}(t) = Q_k(t), S_k^{pos}(t) = S_k(t), \forall k \neq a_t$. The priors of all arms are updated between rounds according to:
\begin{equation}
    Q_k(t+1) = \hat{\lambda} Q_k^{pos}(t) + (1-\hat{\lambda})\hat{\theta}, 
\end{equation}
\begin{equation}
    S_k(t+1) = \hat{\lambda}^2 S_k^{pos}(t)+ \hat{\sigma}_d^2,
\end{equation}
where $\hat{\sigma}_0^2, \hat{\sigma}_d^2, \hat{\theta}, \hat{\lambda}$ denote the subjects' estimates of the parameters in the diffusion process. Notice that in the stochastic 2-armed bandit experiment, since the mean reward for each arm remains constant, we have $\hat{\sigma}_d = \hat{\theta} = 0$, $\hat{\lambda} = 1$ 

The first class of choice models is based on the softmax function with increasing model complexity. 
Coefficient $\beta$ is the \emph{inverse temperature parameter}, which controls the degree of \emph{random exploration}; $\phi$ represents the \emph{uncertainty bonus}, capturing the extent of \emph{directed exploration}; and $\rho$ models \emph{choice perseveration bonus}, reflecting a subject's tendency to repeat previous choices:
\begin{align}
    P(a_t = k) &= \frac{\exp\{\beta Q_k(t) \}}{\sum_{i=1}^K\exp\{\beta Q_i(t)\}},
    \tag{SM-1}\label{eq:SM1}\\
    P(a_t = k) &= \frac{\exp\{\beta[ Q_k(t) + \phi S_k(t)]\}}{\sum_{i=1}^K\exp\{\beta[Q_i(t) + \phi S_i(t)]\}},
    \tag{SM-2}\label{eq:SM2}\\
    P(a_t = k) &= \frac{\exp\{\beta[ Q_k(t) + \phi S_k(t) + \rho \mathbb{I}_{(a_{t-1} = k)}]\}}{\sum_{i=1}^K\exp\{\beta[Q_i(t) + \phi S_i(t) + \rho \mathbb{I}_{(a_{t-1} = i)}]\}}.
    \tag{SM-3}\label{eq:SM3}
\end{align}

We adopt the following interpretation of these parameters from the literature, see, e.g.,  \cite{chakroun2020dopaminergic}:
\begin{itemize}
    \item Higher values of $\beta$ lead to less random and exploratory choices and more ``high-value'' arms;
    \item Higher values of $\phi$ lead to more exploration of uncertain arms, similar to the principle of ``optimism in the face of uncertainty'' in the design of UCB-type algorithms;
    \item Higher values of $\rho$ lead to more repeated choices of the same arm over consecutive periods.
\end{itemize}
To ensure a transparent benchmark, as noted in Section \ref{sec:agents}, we acquire the original experimental data from \citep{chakroun2020dopaminergic}. We re-estimate the model parameters on this dataset to establish a human baseline for comparison against LLMs and algorithms.
The three softmax-based choice rules can be applied to both bandit settings. Additionally, we consider a probit-regression–based choice rule specifically for the 2-armed bandit task \citep{Samuel2019data}, which incorporates the \emph{value difference} $V_t = Q_1(t) - Q_2(t)$, the \emph{total uncertainty} $TU_t = \sqrt{S_1^2(t) + S_2^2(t)}$, and the \emph{relative uncertainty} $RU_t = S_1(t) - S_2(t)$ as covariates.
In each round $t$, the choice probability is given by 
\begin{align}
\small
    P(a_t = 1)=\Phi\left(w_1V_t+w_2RU_t+w_3\frac{V_t}{TU_t}\right),
     \tag{Probit}\label{eq:probit}
\end{align}
     where $\Phi(\cdot)$ is the cumulative distribution function (CDF) of the standard Gaussian distribution with mean 0 and variance 1. 
The parameters have the following interpretation \citep{gershman2018dopaminergic, Samuel2019data}:
\begin{itemize}
    \item Higher values of $w_1$ lead to more frequent choices of empirically higher-value arms, which captures the exploitative behavior of the agent; 
    \item Higher values of $w_2$ lead to more frequent choices of arms with higher uncertainty, which is similar to $\phi$ in the softmax models; 
    \item In addition to $w_1$, $w_3$ is the \emph{uncertainty normalized} value difference which characterizes the tendency to choose empirically high-value arms when the overall uncertainty is low. That is, for a fixed $V_t$, higher values of $w_3$ lead to more frequent choices of high-value arms when the trial is approaching the end and the agent has learned the mean reward of both arms quite accurately. 
\end{itemize}
Similarly, for the 2-armed bandit task, we directly utilize the original human behavioral data from \cite{gershman2018dopaminergic}.
\subsection{Parameter Estimation}
We use a hierarchical Bayesian approach \citep{beharelle2015transcranial, chakroun2020dopaminergic, hBayesDM, ahn2013model} to estimate parameters at the group level for each agent type, while also accounting for individual-level heterogeneity. In the context of behavioral modeling, this approach can yield more robust and interpretable estimates \citep{hBayesDM} compared to traditional maximum likelihood estimation (MLE), which has also been widely used in estimating choice parameters in the previous literature (e.g., \citep{daw2006cortical, bornstein2017reminders, Samuel2019data}). Details of implementation and the hierarchical Bayesian structure could be found in Appendix~\ref{appendix:hierarchical}. 
We note that the hierarchical Bayesian approach estimates the population-level hyper-parameters, such as those controlling the distribution of $\beta$ in \eqref{eq:SM1} among the participants, as well as the individual-level posterior distribution of the parameter for each agent. It thus allows for heterogeneity in the decision-making process across agents.

\subsection{Model Identifiability \& Parameter Recovery}\label{sec:identify}
In this subsection, we discuss the identifiability of the proposed model. This property is crucial as it ensures that the model parameters are uniquely determined from the observed data distribution.
Note that for a choice $a\in [K]$, the softmax-based models defined in Equations (\ref{eq:SM1}, \ref{eq:SM2}, and \ref{eq:SM3}) can be generalized as, \begin{align}
    P_{\theta}(a = k) = \frac{\exp\{\theta^\top I_{k}\}}{\sum_{k'}\exp\{\theta^\top I_{k'}\}}, \tag{General Softmax Choice Model}\label{eq:softmax}
\end{align}
where $\theta \in \mathbb{R}^d$ be the vector of model parameters, and $I_k$ be the index/feature vector associated with the $k$-th arm, e.g., $[Q_k, S_k, \mathbb{I}]^\top$ in the \ref{eq:SM3} model. Similarly, a general form of the \ref{eq:probit} model under the 2-armed bandit can be written as \begin{align}
    P(a=1) = \Phi(\theta^\top I_1). \tag{General Probit Choice Model} \label{eq:general_probit}
\end{align}
Following standard discrete choice theory \citep{train2009discrete}, the inherent translation invariance of the choice models can be resolved by constructing the design matrix using feature differences relative to a base arm. Provided this constructed design matrix maintains full column rank, the global identifiability of the parameters in our sequential setting is guaranteed. For the formal definition and statement regarding model identifiability, see, Appendix \ref{app:id}.

To complement the theoretical results, we conduct a parameter recovery study using synthetic data ($N = 100, T = 300$, 4-armed non-stationary setting). Specifically, each parameter $x$ is drawn from a group-level prior $\mathcal{N}(\mu_x, \sigma_x^2)$, where $\mu_x$ and $\sigma_x$ are calibrated to match our empirical human estimates. We then apply our proposed Bayesian estimation approach to estimate these parameters, confirming that the true underlying parameters are successfully recovered.

Table \ref{tab:recover} summarizes the recovery results. The estimated posterior means for all group-level parameters ($\mu$ and $\sigma$) closely match the true generative values. Furthermore, all true values fall well within their respective 90\% credible intervals, confirming that our Bayesian estimation procedure accurately recovers the underlying parameters without systematic bias.
\begin{table}[ht]
\small
\centering
\caption{Group-Level Parameter Recovery}
\label{tab:recover}
\begin{tabular}{lccc}
\toprule
Parameter & True & Post.\ Mean & 90\% CI \\
\midrule
$\mu_{\beta}$  & 0.168 & 0.171 & $[0.134,\ 0.208]$ \\
$\sigma_{\beta}$ & 0.053 & 0.053 & $[0.038, \ 0.071]$ \\
$\mu_{\phi}$   & 0.879 & 0.872 & $[0.831,\ 0.914]$ \\
$\sigma_{\phi}$ & 0.850 & 0.863 & $[0.741,\ 0.987]$ \\
$\mu_{\rho}$   & 5.450 & 5.412 & $[4.893,\ 5.951]$ \\
$\sigma_{\rho}$ & 0.268 & 0.281 & $[0.214,\ 0.352]$ \\
\bottomrule
\end{tabular}
\end{table}

\subsection{Model Misspecification}\label{sec:mis}
Given the inherent complexity of both the human brain and LLMs, it is infeasible to develop a comprehensive model that fully characterizes their behavior. The proposed choice models in the paper are well-established in the Psychiatry literature  \citep{daw2006cortical,bornstein2017reminders,chakroun2020dopaminergic}, and their behavioral parameters show high consistency with related human cognitive mechanisms.

While our behavioral models are grounded in human cognitive mechanisms, model misspecification may arise when they are applied to non-human agents such as bandit algorithms and LLMs. In this context, we interpret the model coefficients not as an assumption of cognitive equivalence, but rather as a projection framework, i.e., the model projects observed choice trajectories onto an interpretable parameter space, thereby extracting the closest behavioral signatures consistent with the data. 
To illustrate the interpretability of the fitted parameters under model misspecification, we generate synthetic choice data from the bandit algorithms (see Appendix \ref{appendix:algorithm} for detailed algorithms), and fit our behavioral model to the generated trajectories (estimation results in Appendix \ref{appendix:mis}). We further systematically vary the exploration rate ($\epsilon$) of the $\epsilon$-greedy algorithm and the constant of the uncertainty term of the UCB. Re-estimating our model on these adjusted baselines reveals consistent, corresponding trends in the fitted parameters (see Appendix \ref{appendix:mis} for more details). Despite none of these algorithms belonging to our model class, the estimated parameters recover theoretically consistent properties: 
\begin{itemize}
    \item \textbf{Upper Confidence Bound (UCB)}: UCB exhibits large positive estimates of $\phi$, indicating strong directed exploration. This aligns directly with the algorithmic structure of UCB, with explicit uncertainty bonus ${\sigma}_t(k)\cdot\sqrt{\frac{c\log f(t)}{N_t(k)}}$, although the construction is fundamentally different from the uncertainty term in the choice model. Furthermore, comparing within the algorithm reveals that as the exploration hyperparameter $c$ increases, the estimated $\beta$ exhibits a monotone decrease. 
    \item \textbf{Thompson Sampling (TS)}: TS yields small values of $\beta$, reflecting random exploration driven by sampling from posterior distributions.
    \item \textbf{$\epsilon$-greedy}: As the exploration rate $\epsilon$ increases, there is a clear monotonic decline in the inverse temperature parameter ($\beta$), reflecting a higher degree of random exploration. 
\end{itemize}
All three algorithms also exhibit choice perseveration. This occurs because UCB and $\epsilon$-greedy rely on an argmax operation over estimated values, while TS favors the highest posterior mean. Consequently, they repeatedly select the same arm until new observations shift these estimates.
\section{Experimental Results}\label{sec:expresults}
In this section, we present the main empirical findings of our study. First, we compare the out-of-sample predictive performance of various choice models for the agents in both tasks. 
Next, we present and discuss the estimated parameters of the best-performing models. In particular, we focus on comparing the behavioral differences between \emph{thinking and non-thinking settings within the same model architecture.} We also report several model-free metrics to examine the impact of different behaviors on two outcomes of exploitation rate and regret. 

\subsection{Out-of-Sample Performance of Different Choice Models}
To conduct robust model selection based on out-of-sample predictive accuracy, we use a individual-level Bayesian leave-one-out (LOO) cross-validation (CV) approach \citep{vehtari2017practical}. 
For each subject, we compute the log-likelihood of their observed choices using model parameters estimated from the remaining subjects. To avoid the computational burden of refitting the model for each subject, we use Pareto-Smoothed Importance Sampling (PSIS) as an efficient approximation \citep{vehtari2017practical, vehtari2024pareto}. Additional implementation details are provided in Appendix~\ref{appendix:cv}.

Across both experiments, the choice model with the highest score varied across LLMs. 
Even in cases where another model slightly outperforms them, their predictive performance remains highly competitive, making them well-suited for consistent cross-agent comparison.

\subsection{Comparing Behavior Across Agents and Tasks}\label{sec:main_results}
\paragraph{\textbf{Model-based analysis of behavioral parameters.}} To understand how LLMs explore the arms differently from humans, we visualize the estimated parameters of the best-fitted choice model in each experiment. For each population-level mean parameter ($\mu_{n}^x$ in Figure~\ref{fig:hierarchical-model} in Appendix~\ref{appendix:hierarchical}), we visualize the posterior mean and standard deviations using histograms.  We compare these estimates to the corresponding posterior mean for human agents visualized using a dashed horizontal line. To better highlight systematic differences, we categorize models into two groups for comparison. First, we aggregate all basic models (with or without CoT prompting) to assess the behavioral impact of prompt-based thinking capabilities. Second, we compare thinking models with the internal thinking abilities either enabled or disabled (i.e., low vs. high thinking budget or thinking mode toggled on or off). These comparisons allow us to systematically evaluate how enabling thinking influences LLMs’ behavioral parameters across experiments.

\begin{figure}[ht]
    \centering
    \includegraphics[width=0.9\linewidth]{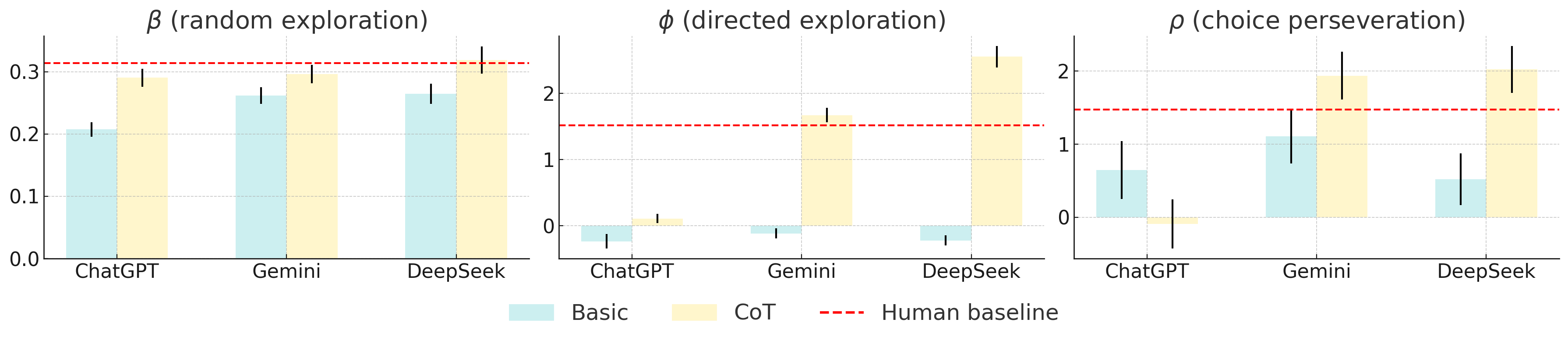}
    \caption{Visualizing the SM-3 model parameters across LLMs in the 2-armed setting}
    \label{fig:basicmodel2arm_SM3pars}
\end{figure}
In the 2-armed bandit setting, figure \ref{fig:basicmodel2arm_SM3pars} visualizes the estimated parameters of the \ref{eq:SM3} model for basic (non-thinking) LLMs. We observe significant differences across basic LLMs and in contrast to humans. Specifically, basic LLMs display higher levels of random exploration ($\beta$), and do not exhibit directed exploration ($\phi$) or choice perseveration ($\rho$) at levels that human agents do. Enabling the thinking through CoT prompting (ask LLMs to "\emph{think out loud}") consistently improves directed exploration and leads to less random exploration, thereby shifting LLMs' exploration behavior closer to human levels. This effect is particularly pronounced for \textsc{Gemini}, which approaches human-like levels of directed exploration and choice consistency. While \textsc{ChatGPT} also benefits from thinking, its directed exploration remains below the human baseline, and does not exhibit choice perseveration. And for \textsc{DeepSeek}, in contrast, shows even higher parameter values of $\phi$ than humans, and a similar trend is observed for choice perseveration. 
\begin{figure}[ht]
    \centering
    \includegraphics[width=0.9\linewidth]{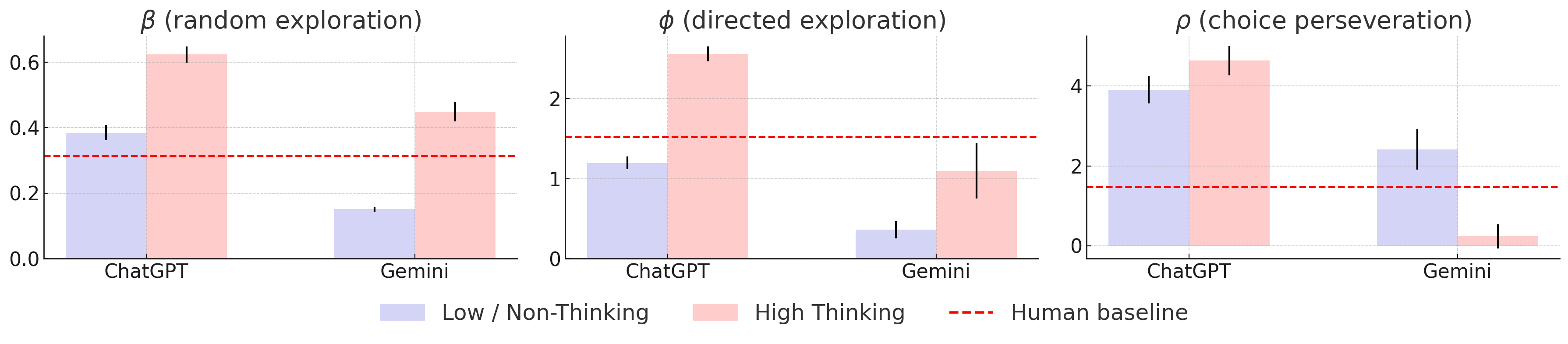}
    \caption{Visualizing the SM-3 model parameters across thinking models in the 2-armed setting}
    \label{fig:thinking2arm_SM3pars}
\end{figure}

To validate the claim that thinking capability increases directed exploration and decreases random exploration, we evaluate this hypothesis under two thinking models (\textsc{ChatGPT-o3-mini} and \textsc{Gemini-2.5-flash}), where the internal thinking trace can be explicitly controlled. Specifically, for \textsc{ChatGPT-o3-mini}, we toggle between low and high thinking modes, and for \textsc{Gemini-2.5-flash}, we manually adjust the thinking budget (with a budget of 0 indicating no thinking). Figure \ref{fig:thinking2arm_SM3pars} visualizes the estimated parameters of the \ref{eq:SM3} model for the two thinking models. 
Consistent with our hypothesis, higher thinking capabilities generally lead to increased directed exploration ($\phi$) and reduced random exploration ($\beta$). Interestingly, for \textsc{ChatGPT-o3-mini} in the high thinking model, the random exploration drops to a level even lower than the human baseline, while the directed exploration exceeds human levels. In contrast, \textsc{Gemini-2.5-flash} shows a more moderate adjustment: increasing the thinking budget raises $\phi$ toward, but not beyond the human baseline, while $\beta$ increases but slightly higher than in humans. 

\begin{figure}[ht]
    \centering
    \includegraphics[width=0.90\linewidth]{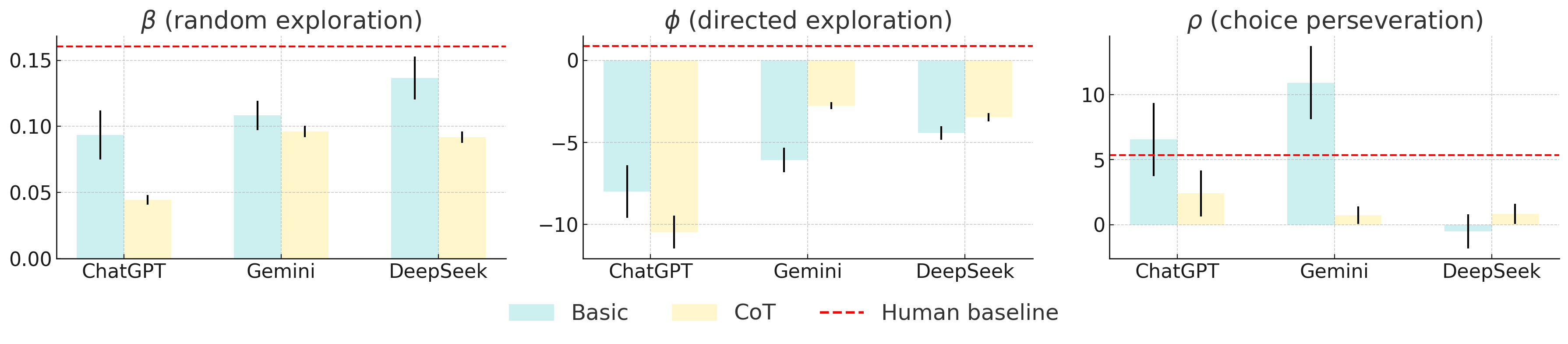}
    \caption{Visualizing the SM-3 model parameters across basic models in the 4-armed setting}
    \label{fig:basic4arm}
\end{figure}
However, in a more complex setting with more arms and a nonstationary environment, LLMs behave markedly differently. Figure \ref{fig:basic4arm} visualizes the estimated parameters of the \ref{eq:SM3} model for basic LLMs in the 4-armed bandit experiment. Across models, basic LLMs consistently exhibit higher levels of random exploration than humans. For \textsc{Gemini} and \textsc{DeepSeek}, enabling thinking capabilities by the CoT prompting increases sensitivity to uncertainty, as reflected by greater values of $\phi$; however, these values remain negative, suggesting that they fail to engage in effective directed exploration. For \textsc{ChatGPT}, thinking does not lead to higher uncertainty sensitivity ($\phi$) or reduced random exploration ($\beta$), suggesting limited benefit from the additional thinking trace in this task. Additionally, we do not observe a consistent pattern in choice perseveration ($\rho$) by LLMs. In contrast, humans continue to engage in choice perseveration. 

\begin{figure}[ht]
    \centering
    \includegraphics[width=0.90\linewidth]{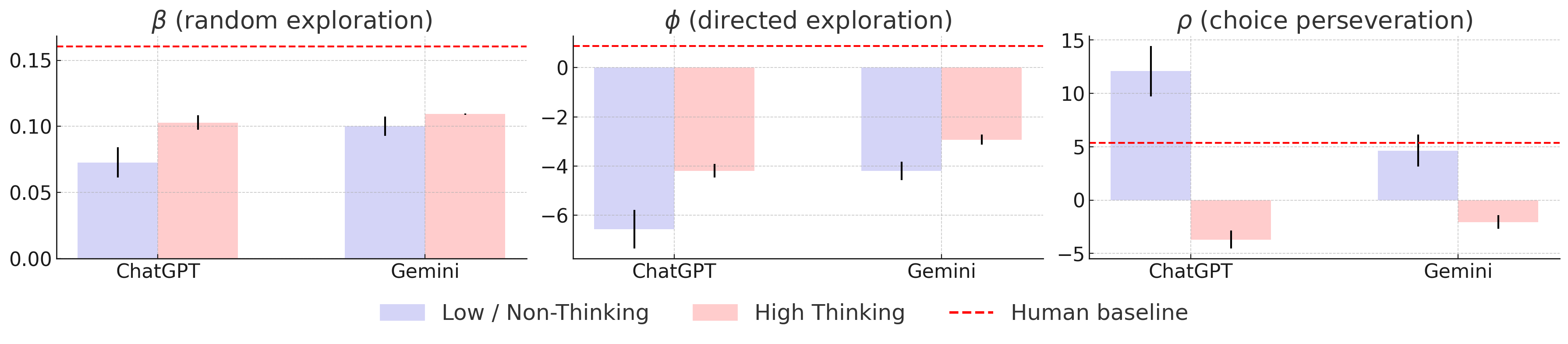}
    \caption{Visualizing the SM-3 model parameters across thinking models in the 4-armed setting}
    \label{fig:thinking4arm}
\end{figure}

Figure \ref{fig:thinking4arm} presents the estimated \ref{eq:SM3} parameters for the thinking models of \textsc{ChatGPT} and \textsc{Gemini} in the 4-armed setting. We observe a similar trend for both models: higher thinking capabilities increase sensitivity to uncertainty and reduce the level of random exploration, but still fail to produce effective directed exploration and consistent pattern of choice perservation.

In conclusion, comparing the parameter estimates in the \ref{eq:SM3} model across the two MAB settings, we find that enabling the thinking trace of LLMs improves alignment with human-like exploration strategies, which combine both random and directed exploration, but the patterns differ by settings. In the 2-armed bandit, thinking capabilities alter LLMs’ exploration toward human-like patterns in terms of reduced random exploration, while in some cases, such as \textsc{ChatGPT}, increasing directed exploration to levels even higher than the human baseline. In contrast, in the more complex, nonstationary 4-armed setting, even with higher thinking budgets, LLMs continue to display substantially lower levels of directed exploration and inconsistent choice perseveration compared to humans. 

Details of the estimated values across different agents (including algorithmic baselines) under choice models in the two MAB experiments are presented in Appendix \ref{appendix:results}. Note that \textsc{DeepSeek-R1} is excluded from the parameter comparison of thinking models, as we cannot vary its internal thinking configuration as in for \textsc{ChatGPT-o3-mini} and \textsc{Gemini-2.5-flash}.

\paragraph{\textbf{Model-free analysis of descriptive metrics.}} 
We next analyze agents' behavior using model-free metrics. Specifically, we focus on two key metrics of \textit{exploitation rate} and \textit{cumulative regret}.

The exploitation rate measures how often an agent selects the arm with the highest observed mean reward. Specifically, we define an action as \emph{exploitative} if the agent chooses the option with the highest estimated mean among all arms at that round. In both experiments, we calculate the average exploitation rate across all participants or replications for each agent type (e.g., humans, LLMs). For both experiments, we report the average exploitation rates across agents.
\begin{table}[ht]
\centering
\caption{Exploitation rates in the 2-armed bandit setting across different agents and baselines.}
\small
\begin{subtable}[t]{0.6\linewidth}
\centering
\caption{Different LLM versions}
\begin{tabular}{lcc|cc}
\toprule
 & \multicolumn{2}{c}{\textbf{Basic Models}} & \multicolumn{2}{c}{\textbf{Thinking Models}} \\
\cmidrule(lr){2-3} \cmidrule(lr){4-5}
\textbf{Agent} & Basic & CoT & Low/Non thinking& High thinking\\
\midrule
\textsc{ChatGPT}   & 0.90 & 0.92 & 0.96 & 0.95 \\
\textsc{Gemini}    & 0.91 & 0.91 & 0.84 & 0.91 \\
\textsc{DeepSeek}  & 0.91 & 0.90 & NA & 0.94 \\
\bottomrule
\end{tabular}
\end{subtable}%
\hfill
\begin{subtable}[t]{0.35\linewidth}
\centering
\caption{Human and algorithms}
\begin{tabular}{lc}
\toprule
\textbf{Agent} & \textbf{Exploitation Rate} \\
\midrule
Human     & 0.89 \\
UCB      & 0.91 \\
TS        & 0.88 \\
$\epsilon$-greedy  & 0.91 \\
\bottomrule
\end{tabular}
\end{subtable}
\label{tab:2armexploitation-rates}
\end{table}
In the 2-armed settings, 
exploitation rates are consistently high across agents and closely match the levels achieved by
algorithmic baselines, as reported in Table \ref{tab:2armexploitation-rates}.

In the 4-armed setting with a longer horizon ($T=300$), we track how agents' exploitation rates evolve by computing them across increasing time windows (e.g., $\tau = $10, 20, 30 rounds, and so on). This helps reveal how different agents adapt their exploitation strategy. To reduce visual clutter, we include only UCB as the representative algorithmic baseline in the main figure. As shown in Figure~\ref{fig:exploit_4arm_all}, LLMs exhibit highly dynamic exploitation behavior in the 4-armed bandit task. Unlike human agents, whose exploitation rate remains relatively stable over time, LLMs show large fluctuations and typically reach a peak around round 50. Interestingly, UCB displays a similar early peak, suggesting that both LLMs and UCB tend to over-commit to certain arms early on, assuming the rewards are fixed. They only begin to adjust after observing enough evidence that the rewards are changing.
This highlights a gap in adaptive capabilities, with humans demonstrating more adaptive adjustments to evolving reward structures. 
\pgfplotsset{compat=1.18}

\usepgfplotslibrary{groupplots,fillbetween}

\tikzset{
  baseLine/.style={blue, thick, solid, mark=*, mark size=1.8pt},
  cotLine/.style={red, thick, dashed, mark=triangle*, mark size=2.2pt},
  ucbLine/.style={brown, thick, dotted, mark=square*, mark size=2.0pt},
  thinkLine/.style={red, thick, dashed, mark=triangle*, mark size=2.2pt},
  lowThinkLine/.style={blue, thick, solid, mark=*, mark size=1.8pt},
  humanLine/.style={black, thick, dashdotdotted, mark=diamond*, mark size=2pt}
}

\begin{figure}
\centering

\begin{tikzpicture}
\begin{groupplot}[
  group style={group size=3 by 1, horizontal sep=18pt, vertical sep=14pt},
  width=0.36\textwidth,
  height=0.26\textwidth,
  grid=both,
  grid style={opacity=0.25},
  xlabel={Rounds  },
  xmin=0
]

\nextgroupplot[ylabel={Exploration (mean)}, title={DeepSeek-V3}]
  \addplot+[mark repeat=5, baseLine, smooth] table[col sep=comma, x=x, y={DeepSeek_mean}] {4_arm_expl.csv};
  \addplot+[mark repeat=5, cotLine, smooth] table[col sep=comma, x=x, y={DeepSeek + COT_mean}] {4_arm_expl.csv};
  \addplot+[mark repeat=5, ucbLine, smooth] table[col sep=comma, x=x, y={UCB_mean}] {4_arm_expl.csv};

\addplot+[mark repeat=5, humanLine, smooth] table[col sep=comma, x=x, y={Human_mean}] {4_arm_expl.csv};
\nextgroupplot[title={ChatGPT-4o-mini}]
  \addplot+[mark repeat=5, baseLine, smooth] table[col sep=comma, x=x, y={ChatGPT_mean}] {4_arm_expl.csv};
  \addplot+[mark repeat=5, cotLine, smooth] table[col sep=comma, x=x, y={ChatGPT + COT_mean}] {4_arm_expl.csv};
  \addplot+[mark repeat=5, ucbLine, smooth] table[col sep=comma, x=x, y={UCB_mean}] {4_arm_expl.csv};
\addplot+[mark repeat=5, humanLine, smooth] table[col sep=comma, x=x, y={Human_mean}] {4_arm_expl.csv};
\nextgroupplot[title={Gemini-2.0-flash}]
  \addplot+[mark repeat=5, baseLine, smooth] table[col sep=comma, x=x, y={Gemini_mean}] {4_arm_expl.csv};
  \addplot+[mark repeat=5, cotLine, smooth] table[col sep=comma, x=x, y={Gemini + COT_mean}] {4_arm_expl.csv};
  \addplot+[mark repeat=5, ucbLine, smooth] table[col sep=comma, x=x, y={UCB_mean}] {4_arm_expl.csv};
\addplot+[mark repeat=5, humanLine, smooth] table[col sep=comma, x=x, y={Human_mean}] {4_arm_expl.csv};
\end{groupplot}
\end{tikzpicture}
\begin{tikzpicture}
    \begin{axis}[
      hide axis,
      xmin=0,xmax=1,ymin=0,ymax=1,
      legend columns=-1,
      legend style={/tikz/every even column/.append style={column sep=1em}, draw=none, fill=none, font=\small}
    ]
      \addlegendimage{baseLine}
      \addlegendentry{Base model}
      \addlegendimage{cotLine}
      \addlegendentry{COT}
      \addlegendimage{ucbLine}
      \addlegendentry{UCB}
      \addlegendimage{humanLine}
      \addlegendentry{Human}
    \end{axis}
  \end{tikzpicture}
\vspace{0.6em}

\begin{tikzpicture}
\begin{groupplot}[
  group style={group size=3 by 1, horizontal sep=18pt, vertical sep=14pt},
  width=0.36\textwidth,
  height=0.26\textwidth,
  grid=both,
  grid style={opacity=0.25},
  xlabel={Rounds  },
  xmin=0
]

\nextgroupplot[ylabel={Exploration (mean)}, title={DeepSeek-R1}]
  \addplot+[mark repeat=5, thinkLine, smooth] table[col sep=comma, x=x, y={\detokenize{DeepSeek (R)_mean}}] {4_arm_expl.csv};
  \addplot+[mark repeat=5, ucbLine, smooth] table[col sep=comma, x=x, y={UCB_mean}] {4_arm_expl.csv};
\addplot+[mark repeat=5, humanLine, smooth] table[col sep=comma, x=x, y={Human_mean}] {4_arm_expl.csv};
\nextgroupplot[title={ChatGPT-o3-mini}]
  \addplot+[mark repeat=5, lowThinkLine, smooth] table[col sep=comma, x=x, y={\detokenize{ChatGPT (R low)_mean}}] {4_arm_expl.csv};
  \addplot+[mark repeat=5, thinkLine, smooth] table[col sep=comma, x=x, y={\detokenize{ChatGPT (R)_mean}}] {4_arm_expl.csv};
  \addplot+[mark repeat=5, ucbLine, smooth] table[col sep=comma, x=x, y={UCB_mean}] {4_arm_expl.csv};
\addplot+[mark repeat=5, humanLine, smooth] table[col sep=comma, x=x, y={Human_mean}] {4_arm_expl.csv};
\nextgroupplot[title={Gemini-2.5-flash}]
  \addplot+[mark repeat=5, lowThinkLine, smooth] table[col sep=comma, x=x, y={\detokenize{Gemini (R low)_mean}}] {4_arm_expl.csv};
  \addplot+[mark repeat=5, thinkLine, smooth] table[col sep=comma, x=x, y={\detokenize{Gemini (R)_mean}}] {4_arm_expl.csv};
  \addplot+[mark repeat=5, ucbLine, smooth] table[col sep=comma, x=x, y={UCB_mean}] {4_arm_expl.csv};
\addplot+[mark repeat=5, humanLine, smooth] table[col sep=comma, x=x, y={Human_mean}] {4_arm_expl.csv};
\end{groupplot}
\end{tikzpicture}
\begin{tikzpicture}
    \begin{axis}[
      hide axis,
      xmin=0,xmax=1,ymin=0,ymax=1,
      legend columns=-1,
      legend style={/tikz/every even column/.append style={column sep=1em}, draw=none, fill=none, font=\small}
    ]
      \addlegendimage{lowThinkLine}
      \addlegendentry{Low thinking}
      \addlegendimage{thinkLine}
      \addlegendentry{Thinking}
      \addlegendimage{ucbLine}
      \addlegendentry{UCB}
      \addlegendimage{humanLine}
      \addlegendentry{Human}
    \end{axis}
  \end{tikzpicture}
\caption{Exploitation rate over time in the 4-armed bandit setting across agents}
    \label{fig:exploit_4arm_all}
\end{figure}

The second metric, cumulative regret, quantifies the performance gap between the agent’s choices and the optimal arm. 
In the 2-armed bandit task, the reward means $\mu_1$ and $\mu_2$ are themselves random variables drawn from a prior distribution at the beginning of each trial. 
This motivates the use of \textbf{\emph{Bayesian regret}}, which averages the performance of a policy over the prior distribution of reward means. 
More precisely, we compute the regret for each agent type across $N$ independent trials. 
In trial $n \in [N]$, a new pair of reward means $\mu_{n,1}, \mu_{n,2}$ is drawn.
Suppose arm $a_{n,t}$ is selected in round $t\in[T]$. 
The \emph{\textbf{empirical Bayesian regret}} up to round $\tau \in [T]$ is computed by averaging over all trials of a given agent type: 
    \begin{align*}
        &\widehat{\mathrm{BayesRegret}}_{\texttt{agent}} (\tau) \nonumber\\
        &= \frac{1}{N}\sum_{n=1}^N\sum_{t=1}^{\tau}\left(\max\{\mu_{n,1}, \mu_{n,2}\} - \mu_{n, a_{n,t}}\right).
    \end{align*}
\input{2arm_regret}
For the 4-armed bandit experiment, recall that the reward sequences follow one of the three pre-defined sets (group 1, 2, and 3). For each agent type (humans, LLMs, algorithms), we average the regret across all $N$ trials. 
More precisely, let $G_n \in \{1,2,3\}$ denote the reward group assignment to trial $n \in [N]$, and $r_{g,k,t}$ denote the reward of arm $k \in [K]$ at round $t\in [T]$ ($K=4, T=300$) in group $g \in \{1,2,3\}$. Let $a_{n,t}$ be the action chosen in trial $n$ at round $t$. The average \textbf{\emph{realized regret}} for a given agent type up to round $\tau \in [T]$ is defined as:
\begin{align*}    &\widehat{\mathrm{RealizedRegret}}_{\texttt{agent}} (\tau) \nonumber\\
&= \frac{1}{N}\sum_{n=1}^N\sum_{t=1}^{\tau}\left(\max_{k\in [K]}r_{G_n,k,t} - r_{G_n,a_{n,t},t}\right).
\end{align*}
\input{4arm_regret}
We emphasize that our performance evaluation is primarily descriptive and comparative rather than prescriptive. Specifically, regret serves as an independent performance metric computed directly from observed choice trajectories, allowing us to evaluate the overall efficacy of exploration-exploitation strategies without directly linking specific estimated behavioral parameters to performance improvements. Comparing the performance across agents and both experiments, as shown in Figure~\ref{fig:regret_2arm}, the cumulative regret remains low across all agents in the 2-armed setting. 
The differences between agents' regrets are roughly equivalent to the regret incurred by a single suboptimal arm choice. 
While algorithms such as TS and $\epsilon$-greedy are known to perform well in the long run, their short-run performance over just 10 rounds is slightly weaker than that of humans and LLMs.
In contrast, we observe that while human agents consistently achieve the lowest regret in the 4-armed bandit experiment, both LLMs and algorithms do much worse (see Figure~\ref{fig:regret_4arm}). Enabling LLMs with thinking traces across three models substantially reduces their regret, bringing their performance closer to human level and surpassing standard algorithms, especially for \textsc{Gemini-2.5-flash} and \textsc{DeepSeek-R1}. These results suggest that thinking capabilities can significantly improve LLMs' E\&E strategies, particularly in complex and non-stationary decision environments.

\subsection{Additional Complementary Analysis}\label{sec:addition}
\begin{figure}[ht]
    \centering
    \includegraphics[width=0.85\linewidth]{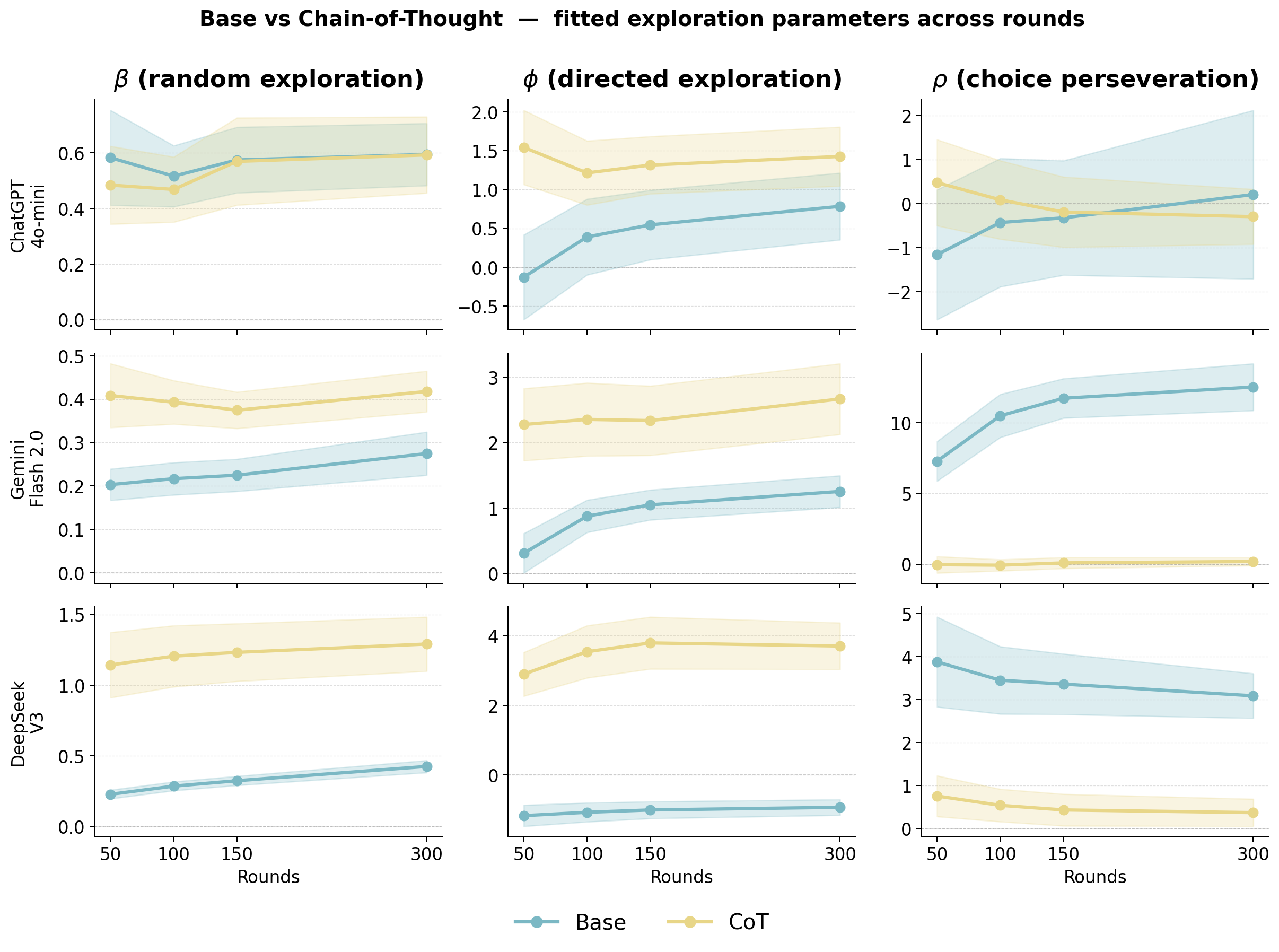}
    \caption{Visualizing the SM-3 model parameters across LLMs in the 2-armed setting (300 rounds)}
    \label{fig:2_arm_300_base}
\end{figure}

\begin{figure}[ht]
    \centering
    \includegraphics[width=0.85\linewidth]{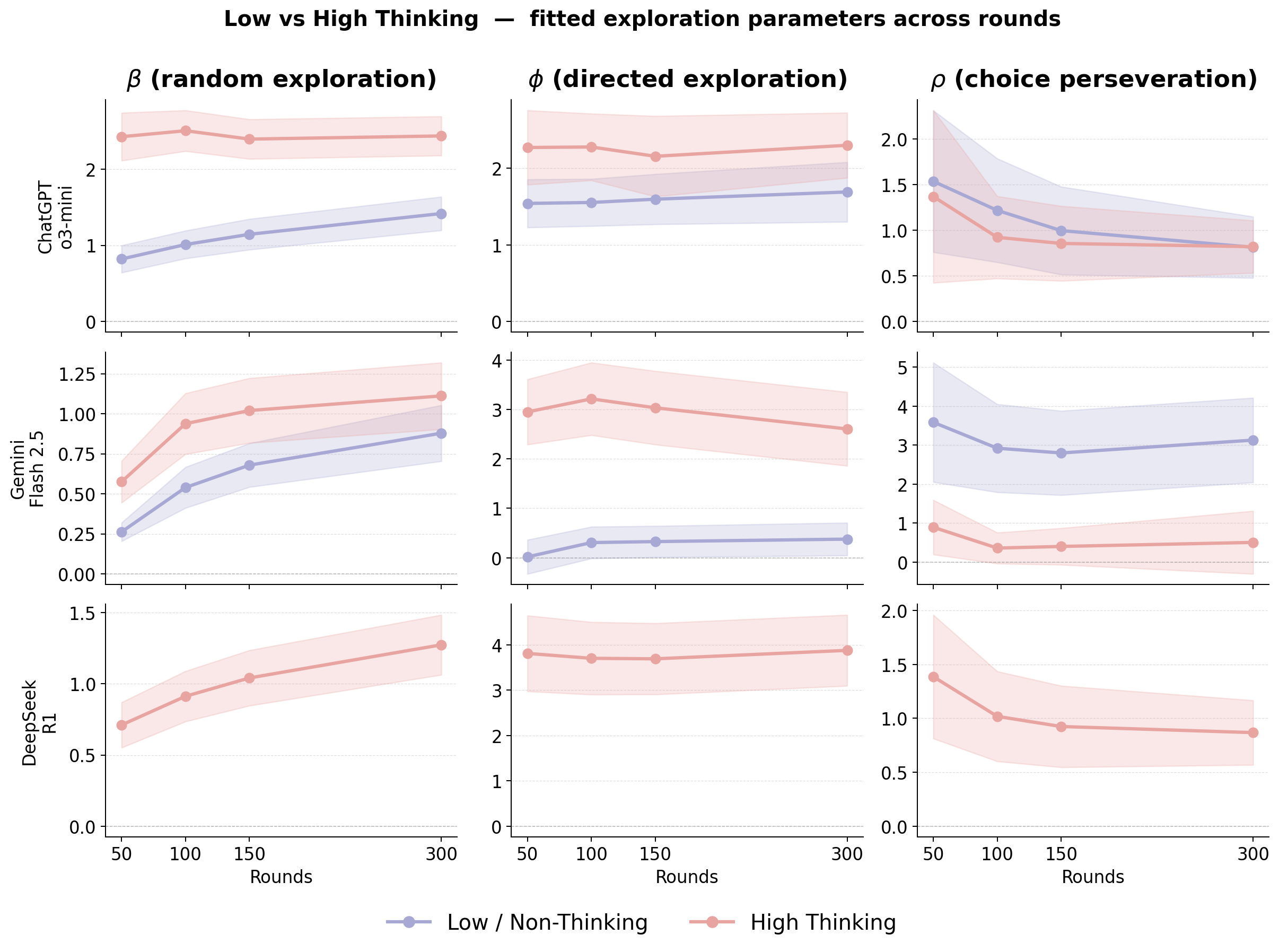}
    \caption{Visualizing the SM-3 model parameters across thinking models in the 2-armed setting (300 rounds)}
    \label{fig:2_arm_300_thinking}
\end{figure}

\begin{figure}[ht]
    \centering
    \includegraphics[width=0.9\linewidth]{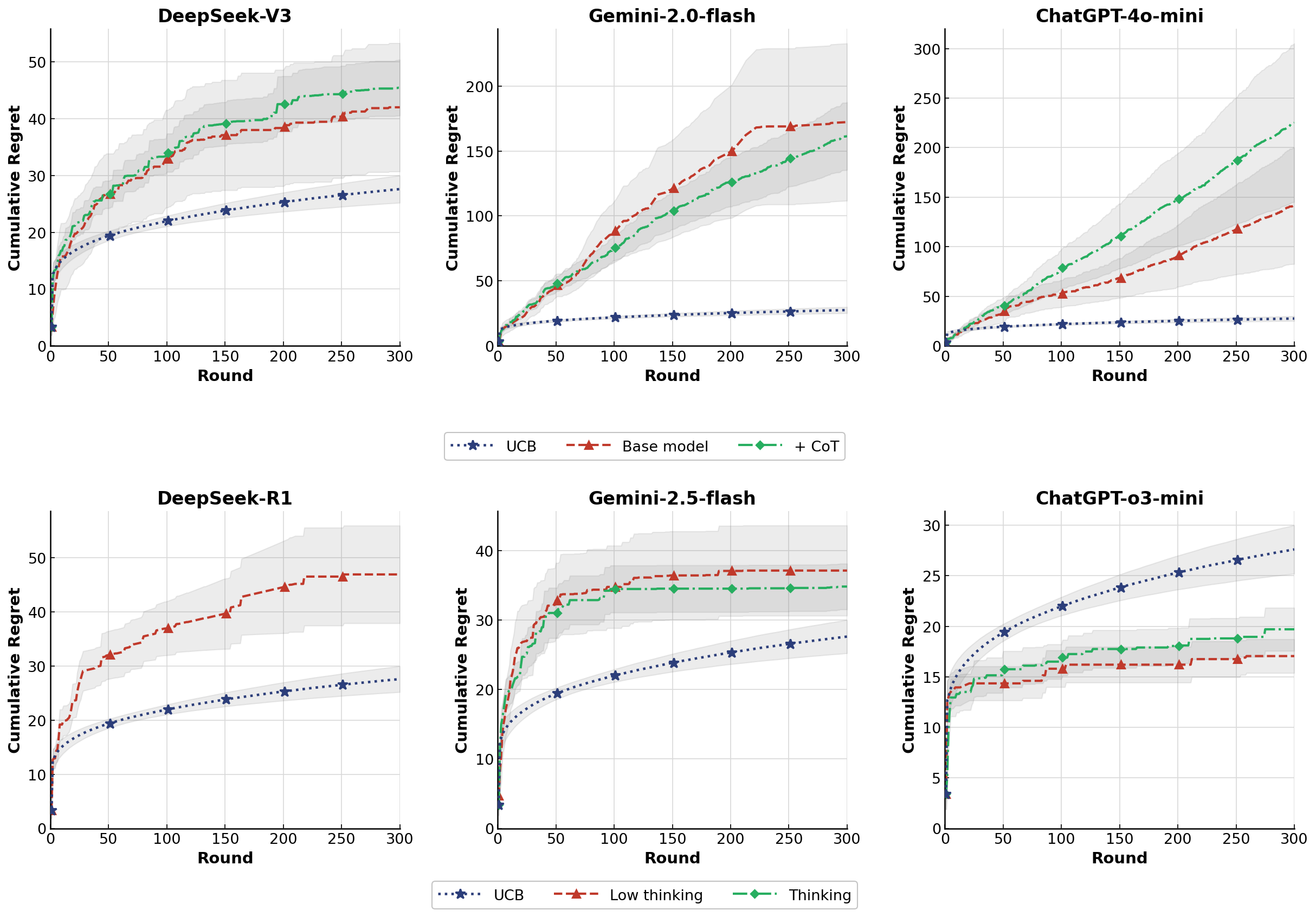}
    \caption{Empirical regret in the 2-armed bandit setting across agents over 300 rounds. Lines represent the average
cumulative regret across trials within a given agent type; shaded areas denote $\pm$1 standard error.}
    \label{fig:regret_2arm_300_rounds}
\end{figure}

In this section, we conduct additional complementary analyses to provide a more comprehensive understanding of the LLMs' exploration-exploitation strategies. Building upon our primary estimation results, we focus on three extensions: evaluating the long-run dynamics of behavioral parameters over an extended horizon under the stationary setting, characterizing the over- and under-exploration behavior, and examining the impact of temperature settings on the inherent exploration behaviors of base models.
\paragraph{\textbf{Long-Run Behavioral Dynamics.}}
Building on our primary structural estimation, we expand the decision rounds to $T=300$ under the same stationary 2-armed bandit setting as the previous analysis, and fit the model at $T = 50, 100, 150, 300$ to investigate long-run exploration behavior and how this behavior evolves over time.
Figures \ref{fig:2_arm_300_base} and \ref{fig:2_arm_300_thinking} indicate a consistent trend aligning with our previous observations: enabling the thinking ability will increase the level of directed exploration while decreasing the random exploration. Meanwhile, tracking these parameters over 300 rounds reveals that LLMs' behavioral parameters exhibit greater variation in the initial phases, but stabilize as the task progresses. Furthermore, we plot the \emph{Empirical Bayesian Regret} across LLMs over 300 rounds. Figure \ref{fig:regret_2arm_300_rounds} demonstrates that enhanced reasoning significantly improves decision-making; specifically, models that utilize advanced reasoning modes achieve a regret level that closely matches the UCB benchmark.

\paragraph{\textbf{Characterizing Over- and Under-Exploration Behavior.}} To complement our primary choice model's focus on exploration types, we further evaluate behavior through the lens of theoretical regret. We adopt the Quantal Choice with Adaptive Reduction of Exploration (QCARE) model \citep{ding2025behavioral}, which defines the arm-pulling score as: \[\theta_{i}(t)=\hat{\mu}_{i}(t)+\frac{1}{(k_{i}(t)+1)^{\alpha}}\epsilon_{it}.\] Here, $\hat{\mu}_{i}(t)$ represents the empirical mean reward of arm $i$ to capture the exploitation component, $k_{i}(t)$ tracks the number of times the arm has been pulled, and $\epsilon_{it}$ introduces independent standard normal shocks to drive exploration. The parameter $\alpha$ quantifies the exploration reduction rate. Specifically, $\alpha=0.5$ represents the optimal trade-off achieving $O(\sqrt{T})$ regret. Any deviation maps directly to asymptotic regret boundaries: $\alpha<0.5$ indicates "over-exploration" causing $\Omega(T^{1-\alpha})$ regret , whereas $\alpha>0.5$ reflects "under-exploration" leading to $\Omega(T^{1-\epsilon})$ regret.

\begin{figure}[ht]
    \centering
    \includegraphics[width=1\linewidth]{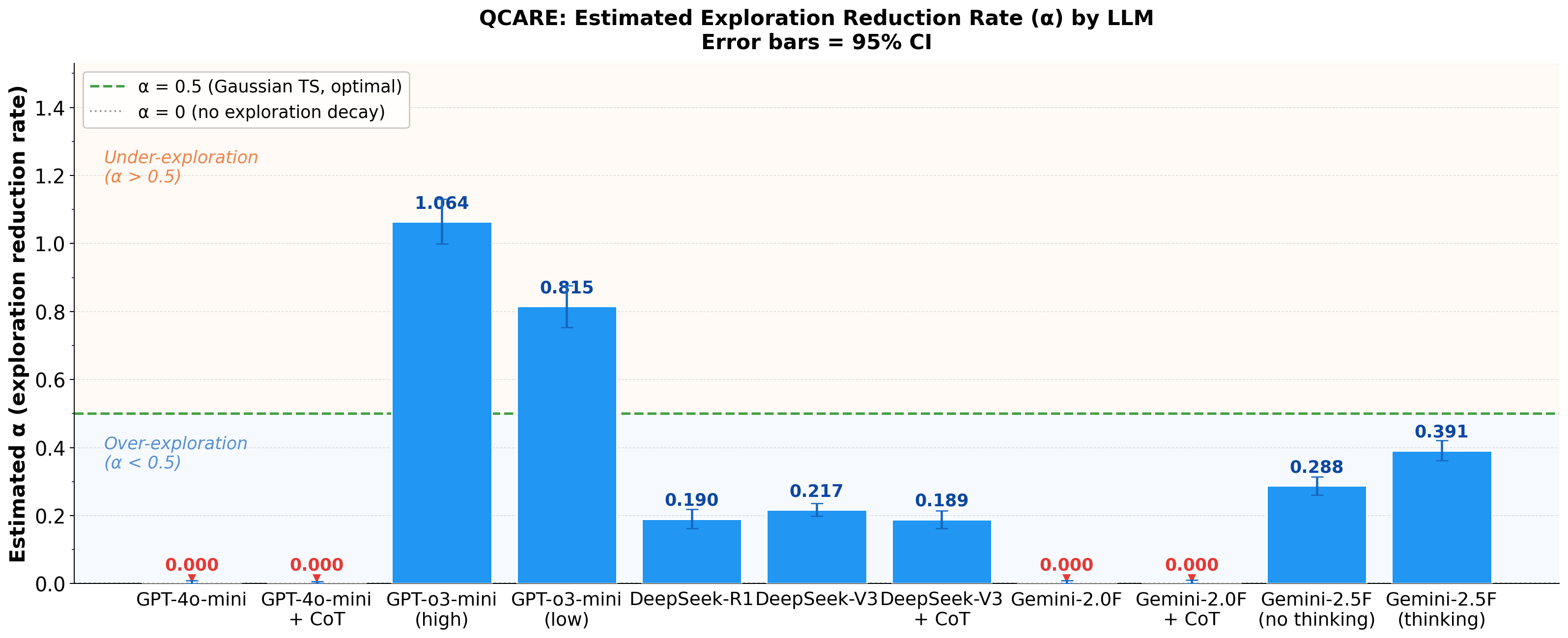}
    \caption{Estimated exploration reduction rates ($\alpha$) across LLMs under the QCARE framework.}
    \label{fig:qcare}
\end{figure}

As illustrated in Figure \ref{fig:qcare}, our estimation shows a notable discrepancy across models in how they navigate this trade-off. Models at the extreme of over-exploration, namely \textsc{GPT-4o-mini} and \textsc{Gemini-2.0-Flash} ($\alpha \approx 0$), exhibit no decay in structural noise over time, a pattern consistent with undirected, \textbf{random} exploration. This overexploration behavior is prevalent across most base models, including the \textsc{DeepSeek} series. At the opposite end, \textsc{GPT-o3-mini} represents an outlier characterized by under-exploration $(\alpha > 0.5$
), where the model prematurely exploits early leading arms, failing to engage in necessary directed exploration. Particularly, \textsc{Gemini-2.5-Flash} with high-thinking mode achieves a higher exploration reduction rate ($\alpha = 0.391$) than its non-thinking mode ($\alpha = 0.288$), making it closest to the theoretically optimal benchmark ($\alpha = 0.5$). This indicates that enabling thinking effectively suppresses persistent structural noise, enabling a more precise calibration between directed exploration and exploitation.

\paragraph{\textbf{Impact of Temperature Setting.}} Finally, to isolate the effect of generation randomness on the exploration-exploitation tradeoff, we conduct an ablation study on the base models across different temperature settings (low, default, and high) while keeping the prompt configuration strictly identical.

\begin{figure}[ht]
    \centering
    \includegraphics[width=1\linewidth]{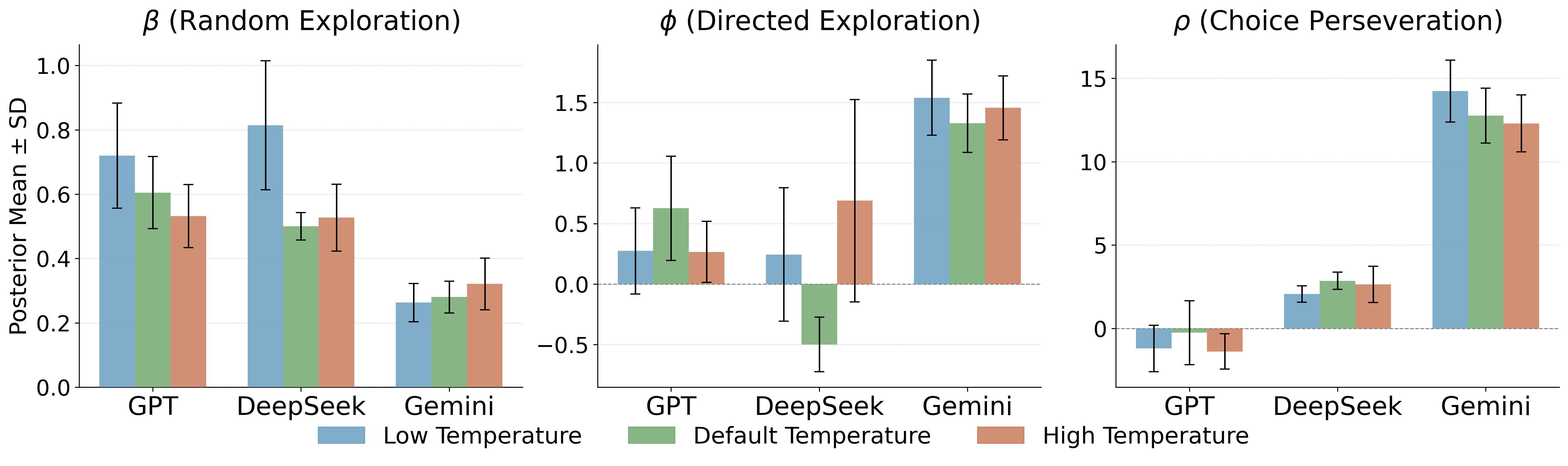}
    \caption{Fitted parameters under varying temperature settings in the stationary 2-armed setting}
    \label{fig:2arm_temp}
\end{figure}

\begin{figure}[ht]
    \centering
    \includegraphics[width=1\linewidth]{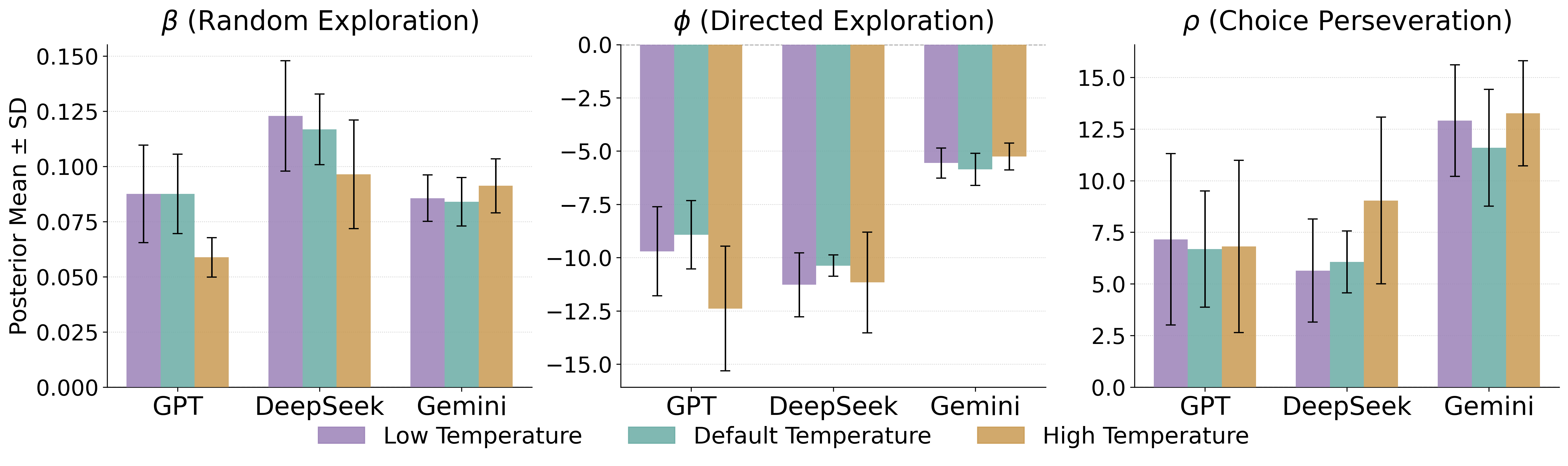}
    \caption{Fitted parameters under varying temperature settings in the non-stationary 4-armed setting}
    \label{fig:4arm_temp}
\end{figure}

As illustrated in Figures \ref{fig:2arm_temp} and \ref{fig:4arm_temp}, our structural estimation reveals while there are slight variations in the fitted posterior means for fitted parameters, these differences are not significant, as evidenced by the heavily overlapping standard deviations. Most importantly, we do not observe any systematic or structural shifts in the models' fundamental exploration strategies simply by altering the temperature. This finding critically reinforces our primary conclusion: the profound improvements in efficient exploration and adaptability observed earlier are genuinely attributable to the models' advanced reasoning and thinking capabilities, rather than being mere artifacts of underlying generation randomness.
\subsection{Key Observations}\label{sec:observation}
Based on the experimental results presented above, in subsection \ref{sec:main_results}, we summarize several key observations regarding LLMs’ E\&E behavior.
\paragraph{\textbf{LLMs exhibit different E\&E behavior compared to humans.}}
The difference is reflected in terms of both the \emph{overall level} of exploration
and, more importantly, the \emph{type} of exploration.
Basic LLMs engage in more random exploration and less direct exploration (Figures \ref{fig:basicmodel2arm_SM3pars}).
Moreover, their exploratory behavior varies drastically over time in the 4-armed bandit experiment (Figure \ref{fig:basic4arm}).
In particular, LLMs tend to explore more at the beginning and less toward the end of the horizon than humans.
In contrast, human agents explore more, use both random and direct exploration strategies, explore at a more consistent level over time, and are more likely to repeat the previous choice.

\paragraph{\textbf{Human agents achieve low regret.}}
Human agents perform well in terms of regret in both experiments (Figures \ref{fig:regret_2arm} and \ref{fig:regret_4arm}), including when we compare them to MAB algorithms that have been shown to have optimal asymptotic regret in stationary settings. We note that our experiments either have a short horizon (2-armed bandit) or are non-stationary (4-armed bandit), and hence do not satisfy the assumptions under which the performance guarantees of these algorithms are established.
Nevertheless, our results reveal that human agents, just relying on their intuitions, can navigate decision-making in unknown and complex environments remarkably well. 

\paragraph{\textbf{Enabling thinking increases direct exploration and reduces regret for LLMs.}}
Producing thinking traces before a final response systematically changes the E\&E behavior of LLMs.
In particular, they increase the extent to which LLMs engage in direct exploration (e.g., Figures \ref{fig:basicmodel2arm_SM3pars}, \ref{fig:thinking2arm_SM3pars}).
 Interestingly, we observe that in the stationary setting, LLMs with thinking trace can behave similarly to humans, especially \textsc{Gemini}. In the non-stationary setting, however, even thinking models exhibit higher levels of random exploration than humans and almost no directed exploration. Moreover, the thinking capability substantially improves the performance of LLMs measured in terms of regret.

\section{Conclusions and Future Work}
This work examines how large language models (LLMs) make decisions under uncertainty by analyzing their exploration–exploitation (E\&E) behaviors in multi-armed bandit (MAB) experiments. Using interpretable behavioral models from cognitive science and psychology, we analyze how thinking capabilities impact the internal strategies and performance of LLMs. We compare these behaviors against human participants and algorithmic baselines across both simple and complex decision-making environments.

For the implications for using LLMs in dynamic decision-making tasks,
first, our results highlight the limitations of LLMs in simulating the E\&E behavior of humans, particularly in complex, non-stationary settings. Specifically, basic LLMs exhibit random and uneven exploration over time, lacking the consistency, directed exploration, and repetition biases characteristic of human behavior. Although augmenting LLMs with thinking traces steers their behavior toward human-like E\&E strategies, we continue to observe systematic deviations. The results suggest that embedding explicit prompts for uncertainty estimation and exploration reasoning could help align the behavior of LLMs with human strategies. 

Importantly, these findings also have implications beyond our controlled experimental settings. Many real-world scenarios, for example, scientific discovery (see \citet{ren2025towards} for a recent survey on LLMs-based scientific agents), require agents to perform effective exploration that balances novelty-seeking with goal-directed efficiency. In such contexts, lack of consistent directed exploration observed in LLMs could lead to missed opportunities for discovering high-value solutions or pursuing promising hypotheses. Our results further suggest that enabling internal thinking capabilities or incorporating explicit prompt designs may help LLM-based agents explore more actively and robustly, thereby improving their potential for complex, open-ended problem-solving without requiring fundamentally new architecture.

In addition, when used for automated decision-making, we find that basic LLMs achieve higher regret levels compared to humans in complex non-stationary tasks. Notably, this can be largely mitigated when enabling their internal thinking capabilities. This suggests that targeted prompt design could elicit more effective exploration strategies and improve performance, even without modifying model architecture.

While our work reveals systematic differences between LLMs and human E\&E strategies in MAB experiments, we do not directly address how to reduce these discrepancies. In particular, we do not propose specific methods to align LLM behavior more closely with human strategies. Future work could explore techniques such as prompt engineering, instruction fine-tuning, or reinforcement learning to guide LLMs toward more adaptive, human-like exploration policies.
Another key limitation of our current study is that our human dataset does not include demographic features (e.g., age, sex, education level, race), which could influence decision-making strategies. Consequently, our LLM evaluation does not account for such individual-level variability, and we do not incorporate such demographic attributes into LLM prompts. Future work could explore how incorporating such information impacts LLMs' behavior and their alignment with human strategies. 
%
%
%




\bibliographystyle{informs2014} 
\bibliography{ref} 

\begin{thebibliography}{3}
\providecommand{\natexlab}[1]{#1}
\providecommand{\url}[1]{\texttt{#1}}
\providecommand{\urlprefix}{URL }

\bibitem[{Smith(2005)}]{smith2005}
Smith J (2005) Optimal resource allocation in humanitarian logistics.
  \emph{Journal of Operations Research} 30(2):123--135.
  
\bibitem[{Jones(2010)}]{jones2010}
Jones S (2010) Stochastic programming models for humanitarian logistics.
  \emph{INFORMS Mathematics of Operations Research} 35(4):567--580.

\bibitem[{Brown(2015)}]{brown2015}
Brown D (2015) \emph{Introduction to Stochastic Programming} (Springer).

\end{thebibliography}

\newpage
\begin{APPENDICES}

\section{Experimental Setup Details}\label{appendix:setup}
\subsection{Human Data}\label{appendix:human}

In this section, we summarize the instructions provided to human participants in both bandit tasks. These instructions are adapted directly from the original studies \citep{Samuel2019data, chakroun2020dopaminergic, daw2006cortical} and used to form the prompts given to LLMs. Our goal is to ensure that human and LLM agents are presented with comparable information, such as task structure and task objectives, therefore enabling a fair comparison of their decision-making behavior.

\paragraph{2-armed bandit.} As reported in \cite{Samuel2019data}, each of the 45 participants play 20 independent games with each game consisting of 10 rounds and a distinct pair of slot machines. In each round, participants choose between two colored buttons representing slot machines and receive feedback in the form of integer rewards. The instructions also emphasized that the rewards are stochastic, and participants will win or lose points based on their choice. Each participant is paid a fixed amount \$1.5. 

\paragraph{4-armed bandit.} Data corresponds to a double-blind, counterbalanced, placebo-controlled within-subjects study \cite{chakroun2020dopaminergic}. Participants perform a restless four-armed bandit task under three drug conditions, including placebo. In our analysis, we exclusively use data obtained under the placebo, corresponding to 31 male participants who were mainly university students.
All participants underwent a medical screening prior to the experiment, which included an electrocardiogram (ECG) and a health interview.

Each participant completes 300 rounds of a decision-making task involving four options, grouped into four blocks of 75 rounds. At each round, participants select one of four colored squares displayed on a screen, each representing a different \textit{‘bandit’}. After making a choice, they receive feedback indicating the number of points earned at that round.

Participants are informed that the goal is to maximize their total number of points throughout the task. The reward associated with each bandit drifts stochastically over time, following an independent Gaussian random walk. Importantly, participants are told that the payout would be tied to their total accumulated points, with 5 cents paid out for every 100 points earned. Specifically, they receive a fixed compensation of €270 and an additional performance-based bonus (€30–50). The full distribution of reward drift is not revealed, so participants must continuously track value estimates and adjust their exploration and exploitation strategies accordingly.
\newpage
\subsection{Algorithmic Baseline Setup}\label{appendix:algorithm}
We implemented three standard algorithmic baselines for the MAB tasks: Upper Confidence Bound (UCB), Thompson Sampling (TS), and $\epsilon$-greedy. These algorithms serve as interpretable performance benchmarks for comparison with human and LLM agents.
\paragraph{Upper Confidence Bound (UCB).} In the $K$-armed bandit task, we first choose each arm once, and subsequently choose arm $a_t$ on each round $t\in [T]$ according to: \begin{equation}\label{eq:}
    a_t = \arg\max_{k\in [K]}\left(\hat{\mu}_t(k) + \hat{\sigma}_t(k)\cdot\sqrt{\frac{2\log f(t)}{N_t(k)}}\right),
\end{equation}
where,\begin{itemize}
    \item $f(t) = 1+t\log^2(t)$;
    \item $\hat{\mu}_t(k), \hat{\sigma}_t(k)$ are the sample mean and sample standard deviation of rewards for arm $k$ up to round $t$, respectively;
    \item $N_t(k)$ is the number of times arm $k$ has been selected up to round $t$.
\end{itemize}
For initialization, when an arm has fewer than two samples, we set its sample deviation to a fixed piror value. Specifically, we set $\hat{\sigma}_t(k) = \sqrt{10}$ for the 2-armed bandit task and $\hat{\sigma}_t(k) = 2$ for then 4-armed bandit task for every arm $k\in [K]$.

This formulation adapts the classic asymptotically optimal UCB algorithm described in \citep{lattimore2020bandit} by scaling the exploration term with the empirical standard deviation, thereby allowing the confidence interval to adjust based on the observed variability of each arm. 

\paragraph{Thompson Sampling (TS).}
We adopt a Bayesian formulation with a Normal-Inverse-Gamma conjugate prior to model rewards with unknown variance. Assume rewards for each arm $k$ are normally distributed:
\begin{equation}
    r_t(k) \sim \mathcal{N}(\mu_k, \sigma_k^2)
\end{equation}
The prior over $(\mu_k, \sigma_k^2)$ is given by:
\begin{equation}
    \mu_k \mid \sigma_k^2 \sim \mathcal{N}(\mu_0, \sigma_k^2 / \lambda_0), \quad \sigma_k^2 \sim \text{Inv-Gamma}(\alpha_0, \beta_0)
\end{equation}
At each round $t$, the algorithm samples from the posterior distribution of each arm and selects the arm with the highest sampled mean:
\begin{enumerate}
    \item For each arm $k\in [K]$, sample:
    \[
    \sigma_k^2 \sim \text{Inv-Gamma}(\alpha_k, \beta_k), \quad
    \tilde{\mu}_k \sim \mathcal{N}(\mu_k, \sigma_k^2 / \lambda_k)
    \]
    \item Select arm:
    \[
    a_t = \arg\max_{k\in[K]} \tilde{\mu}_k
    \]
    \item Observe reward $r_t$, and update posterior for arm $a_t$ as follows: Let the current parameters for the selected arm be $(\mu_{a_t}, \lambda_{a_t}, \alpha_{a_t}, \beta_{a_t})$. The posterior update is:
\[
\begin{aligned}
\lambda_{a_t} &\leftarrow \lambda_{a_t} + 1 \\
\mu_{a_t} &\leftarrow \frac{\lambda_{a_t} \mu_{a_t} + r_t}{\lambda_{a_t} + 1} \\
\alpha_{a_t} &\leftarrow \alpha_{a_t} + \frac{1}{2} \\
\beta_{a_t} &\leftarrow \beta_{a_t} + \frac{(r_t - \mu_{a_t})^2 \cdot \lambda_{a_t}}{2(\lambda_{a_t} + 1)}
\end{aligned}
\]
\end{enumerate}
We use weakly informative priors in the 2-armed bandit task: $\mu_0 = 0$, $\lambda_0 = 1$, $\alpha_0 = 1$, $\beta_0 = 1$. And specifically, for the 4-armed bandit task, we set $\mu_0 = 50$, and other hyperparameters remain the same as in the 2-armed task.

\paragraph{$\epsilon$-greedy.} For the $\epsilon$-greedy algorithm, the agent selects the arm with the highest empirical mean with probability $1-\epsilon$, and explores a random arm with probability $\epsilon$:\begin{equation}
    a_t =
\begin{cases}
\arg\max_{k\in [K]} \hat{\mu}_t(k), & \text{with probability } 1 - \varepsilon \\
\text{random arm}, & \text{with probability } \varepsilon
\end{cases}
\end{equation}

We perform a grid search over $\epsilon \in \{0.1,0.2,\dots,0.9\}$ and find that $\epsilon = 0.1$ yields the lowest average regret across trials, so we adopt this value in all reported results.

To ensure that all algorithms start with well-defined estimates and avoid degenerate behavior in early rounds, we initialize each agent by pulling every arm once before applying the main decision rule. This warm-up phase ensures that all arms have at least one reward observation, which is particularly important for algorithms such as UCB (to avoid division by zero) and $\epsilon$-greedy (to compute initial sample means). Although Thompson Sampling with a prior does not strictly require warm-up, we adopt the same initialization procedure for consistency across all baselines.

\subsection{Prompt Design}\label{appendix:prompt}
\newtcolorbox{promptbox}[2][]{%
  colback=white,
  colframe=blue!75!black,
  fonttitle=\bfseries\normalsize,
  title={Prompt: #2},
  coltitle=white,
  colbacktitle=blue!75!black,
  boxrule=0.8pt,
  arc=1mm,
  top=1mm, bottom=1mm, left=1mm, right=1mm,
  enhanced,
  #1
}

\begin{promptbox}{2-armed Bandit Experiment}
\textcolor{blue!75!black}{\textbf{System:}} You are a real human agent playing with two slot machines, labeled 1 and 2, which provide uncertain rewards over time. You will play 20 games, each with a different pair of slot machines. Each game consists of 10 rounds. In each round, you are asked to choose one machine to play, and you will win or lose points based on your choice. Your objective is to maximize your total reward.

\medskip
\textcolor{blue!75!black}{\textbf{User:}}

\textbf{If} round == 1:  
 
You are now performing game: \{block\_index\}, round 1.

\textbf{Else}:  
 
You are now performing game: \{block\_index\}, round: \{round\_index\}.  
Your history is provided below, which includes the “choice” you made and the corresponding “reward” you received in each round. Negative reward means losing points, and positive means winning points. \{JSON\}

\medskip
\textbf{Final instruction:}

– V1: Which machine do you choose between machines 1 and 2?  
You can think out loud and answer the number.  

– V2: Which machine do you choose between machines 1 and 2?  
Do not explain, answer the number.
\end{promptbox}

\newtcolorbox{promptboxfour}[2][]{%
  colback=white,
  colframe=green!70!black,
  fonttitle=\bfseries\normalsize,
  title={Prompt: #2},
  coltitle=white,
  colbacktitle=green!70!black,
  boxrule=0.8pt,
  arc=1mm,
  top=1mm, bottom=1mm, left=1mm, right=1mm,
  enhanced,
  #1
}

\begin{promptboxfour}{4-armed Bandit Experiment}
\textcolor{green!70!black}{\textbf{System:}} You are a real human agent playing with four slot machines, labeled 0, 1, 2, and 3, which provide uncertain rewards over time. You will play a single game consisting of 300 rounds. In each round, you choose one machine to play and receive points based on your choice. Your objective is to maximize your total reward throughout the experiment.

\medskip
\textcolor{green!70!black}{\textbf{User:}}

\textbf{If} round == 1:  

You are now performing round 1.

\textbf{Else}:  

You are now performing round: \{round\_index\}.  
Your history is provided below, which contains the “choice” you made and the corresponding “reward” you received in each “round.”  
\{JSON\}

\medskip
\textbf{Final instruction:}

– V1: Which machine do you choose between machines 0, 1, 2 and 3?  
You can think out loud and answer the number.

– V2: Which machine do you choose between machines 0, 1, 2 and 3?  
Do not explain, answer the number.
\end{promptboxfour}
\paragraph{An example of JSON format.}Figure~\ref{fig:json_prompt} shows an example of the JSON-formatted prompt shown to the LLM, which includes the choice and reward history from previous rounds:

\begin{figure}[ht]
    \centering
    \begin{minipage}{0.9\linewidth}
    \begin{lstlisting}[basicstyle=\ttfamily\small]
[
  {
    "round": 1,
    "choice": 2,
    "reward": 45
  },
  {
    "round": 2,
    "choice": 1,
    "reward": 21
  }
]
    \end{lstlisting}
    \end{minipage}
    \caption{Example JSON-format prompt}
    \label{fig:json_prompt}
\end{figure}
\paragraph{Examples of LLMs' responses} To verify that our \textit{"Think out loud"} prompt encourages LLMs to elicit a reasoning process similar to CoT prompting, we provide several illustrative examples of LLMs' responses.

\newtcolorbox{responsebox}[3][]{
  colback=white,        
  colframe=black,       
  fonttitle=\bfseries\normalsize, 
  title={#2: #3},    
  coltitle=white,       
  colbacktitle=black,
  boxrule=0.8pt,        
  sharp corners,        
  top=1mm, bottom=1mm, left=1mm, right=1mm, 
  enhanced,             
  attach boxed title to top left={yshift=-2mm, xshift=10mm},
  boxed title style={sharp corners, boxrule=0pt}, 
  #1 
}

\begin{responsebox}{2-armed bandit experiment response}{\textsc{ChatGPT}}
\vspace{0.5em}
Based on the history provided:

\begin{itemize}
    \item In round 1, I chose machine 1 and received a reward of 21.
    \item In round 2, I chose machine 2 and received a reward of 6.
    \item In round 3, I chose machine 1 again and received a reward of 24.
\end{itemize}

So far, I have played machine 1 twice with good rewards (21 and 24 points) and machine 2 once with a lower reward (6 points). This suggests that machine 1 has been performing better for me based on the previous rounds.

\medskip

I have a strong inclination to choose machine 1 again for this round, as it seems to have a higher average reward based on the history so far.

\medskip

Therefore, I will choose: Arm 1

\end{responsebox}

\begin{responsebox}{2-armed bandit experiment response}{\textsc{Gemini}}
\vspace{0.5em}
Machine 1 has yielded -8 and -6, for an average of -7.

Machine 2 has yielded -8, 0, and -9, for an average of -5.67.

\medskip
While Machine 2's average is slightly better, the small sample size makes it hard to be confident. Machine 2 also has shown more variability, having given a reward of 0 once.  Since I've played machine 1 less, I'll try it again to get a better sense of its average reward.
\end{responsebox}

\begin{responsebox}{2-armed bandit experiment response}{\textsc{DeepSeek}}
\vspace{0.5em}
Let me analyze the history:
\medskip
\begin{itemize}
    \item Machine 1 was chosen twice, yielding rewards of 5 and 12. The average reward for Machine 1 is (5 + 12) / 2 = 8.5.
    \item Machine 2 was chosen once, yielding a reward of 10. The average reward for Machine 2 is 10.
\end{itemize}
Based on the average rewards, Machine 2 has a higher average reward (10) compared to Machine 1 (8.5). Therefore, I will choose Machine 2 to maximize my expected reward.
\end{responsebox}
\newpage
\section{Estimation and Model Selection}\label{appendix:model}
In this section, we provide details on the hierarchical Bayesian structure for parameter estimation and the cross-validation approach for model selection.

\subsection{Parameters Estimation \& Hierachical Bayesian Structure}\label{appendix:hierarchical}

\begin{wrapfigure}{r}{0.35\textwidth}
\centering
\begin{tikzpicture}[
  node distance=10mm and 12mm,
  every node/.style={font=\small},
  latent/.style={circle,draw,minimum size=10mm},
  obs/.style={rectangle,draw,fill=gray!30,minimum size=8mm},
  plate/.style={draw, rectangle, rounded corners, inner sep=5pt},
  factor/.style={circle,draw, double, minimum size=6mm}
]

\node[latent] (M) at (2,0) {\(\mu^x_n\)};
\node[latent, right=of M, xshift=-3mm] (L) {\(\sigma^x_n\)};

\node[latent, below =of M, xshift=10mm, yshift = 5mm] (x) {\(x_{n,s}\)};
\draw[->] (M) -- (x);
\draw[->] (L) -- (x);

\node[factor, below=of x, yshift=5mm] (P) {\(P_{n,s,t}\)};
\draw[->] (x) -- (P);

\node[plate, fit=(P), name=plateT] {};

\node[plate, fit=(x)(plateT), name=plateS] {};

\node[plate, fit=(M)(L)(plateS), name=plateD] {};
\end{tikzpicture}
\caption{Hierachical Bayeisan Structure}
\label{fig:hierarchical-model}
\end{wrapfigure}

Following \citep{chakroun2020dopaminergic, Samuel2019data}, in the 4-armed bandit task, we fix $\hat{\sigma}_0^2 = 4, \hat{\sigma}_d^2 = 2.8, \hat{\lambda} = 0.9836, \hat{\theta} = 50$ to be the same values as the true parameters in the diffusion process. We set $Q(1) = 50$ at the start of each independent trial. Similarly, in the 2-armed bandit task, we set $\hat{\sigma}^2_0 = 10$, and $Q(1) = 0, S(1) = 10$. These settings enable direct comparison with previous studies.

As shown in Figure~\ref{fig:hierarchical-model}, the model adopts a hierarchical structure in which any model parameter can be expressed in the form \(x_{n,s}\). In other words, each parameter---whether it is $\beta$, $\phi$, $\rho$ in SM-based choice models or $w_1$, $w_2, w_3$ in the Probit regression model---is modeled in this generic form. Specifically, for each subject $s$ in group $n$, the individual-level parameter $x_{n,s}$ is assumed to be drawn from a population-level normal prior distribution:
\begin{equation}
    x_{n,s} \sim N(\mu_n^x, (\sigma_n^x)^2).
\end{equation}
Here, each group corresponds to a specific category of agents (e.g., humans or LLMs), with each subject representing an individual human or a replication in an LLM experiment. The population-level mean $\mu_n^x$ is assigned a non-informative uniform prior over the interval $[x_{\min}, x_{\max}]$, while the population-level standard deviation $\sigma_n^x$ is given a half-Cauchy prior, i.e. 
\begin{equation}
    \mu_n^x \sim \operatorname{Uniform}(x_{\min}, x_{\max}), \quad \sigma_n^x \sim \operatorname{half\text{-}Cauchy}(0, 1).
\end{equation}

Specifically, in the softmax model, we set $\beta_{\min} = 0, \beta_{\max} = 10$, and $\phi_{\min} = \rho_{\min} = -\infty, \phi_{\max} = \rho_{\max} = \infty$, which corresponds to assigning non-informative priors on $\phi$ and $\rho$. And in the probit model, we set ${w_1}_{\min} = 0, {w_2}_{\min} = {w_3}_{\min} = -1$ and ${w_1}_{\max} = {w_2}_{\max} = {w_3}_{\max} = 5$.

We aim to infer the posterior distribution over all model parameters given the observed choice data. We then use Markov Chain Monte Carlo (MCMC) sampling to approximate the posterior. We implement this procedure using PyStan package, which interfaces with the Stan probabilistic programming framework \citep{carpenter2017stan}. Each model is fitted using four independent chains, each with 1000 iterations and 500 warm-up steps. Posterior summaries are computed from the combined post-warm-up samples.

\subsection{ Bayesian
leave-one-out (LOO) cross-validation (CV)}\label {appendix:cv}
We implement a individual-level Bayesian leave-one-out (LOO) cross-validation procedure to evaluate the out-of-sample predictive performance of each choice model. Specifically, we iteratively exclude one subject as the validation set, fit the model to the remaining subjects using MCMC sampling, and compute the predictive log-likelihood of the held-out subject’s choices. The resulting log-likelihood serves as the individual-level LOO-CV score, with higher scores indicating better predictive accuracy.

As full MCMC refitting for each held-out subject is computationally intensive, we adopt Pareto-Smoothed Importance Sampling (PSIS) \citep{vehtari2024pareto} as an efficient approximation to exact LOO-CV. For each participant, we compute the sum of pointwise log-likelihoods across all rounds and trials based on posterior samples, and apply importance sampling to these individual-level likelihoods to estimate the leave-one-out predictive density.
We implement this procedure using the \texttt{loo} function in the ArviZ package, applied to posterior draws generated from PyStan.

\section{Model Identifiability}\label{app:id}
\begin{definition} (Identifiability) 
Let $\mathcal{P} = \{P_\theta: \theta \in \Theta \}$ be a choice model with parameter space $\Theta$. The model $\mathcal{P}$ is identifiable if the mapping $\theta \to P_{\theta}$ is injective. That is, for all $\theta_1, \theta_2 \in \Theta$, \begin{align*}
    &\forall k \in [K], t \in [T], P_{\theta_1}( a_t = k\mid I_{1,t},\dots,I_{K,t}) = P_{\theta_2} (a_t = k\mid I_{1,t},\dots,I_{K,t}) \\
    &\implies \theta_1 = \theta_2, \forall \theta_1, \theta_2 \in \Theta
\end{align*}
\end{definition}
In our sequential setting, the full likelihood of the observed history is the product of the conditional choice probabilities at each step $t$. Because the decisions are conditionally independent given the feature vectors $I_{k,t}$, identifiability at the step-level is logically sufficient to guarantee the identifiability of the entire sequential data-generating process.
\begin{proposition}\label{prop:id}
Let the feature vector for arm $k$ at time $t$ be $I_{k,t} \in \mathbb{R}^d$. The \ref{eq:softmax} and \ref{eq:general_probit} parameters $\theta$ are identifiable if the design matrix $\mathbf{X}_{1:T} = [\mathbf{X}_1^\top, \dots, \mathbf{X}_T^\top]^\top \in \mathbb{R}^{T(K-1) \times d}$ has full column rank $d$, where each $\mathbf{X}_t$ is defined as:
\[
\mathbf{X}_t := \begin{bmatrix} 
(I_{1,t} - I_{K,t})^\top \\ 
\vdots \\ 
(I_{K-1,t} - I_{K,t})^\top 
\end{bmatrix} \in \mathbb{R}^{(K-1) \times d}.
\]
\end{proposition}
The construction of the design matrix $\mathbf{X}_t$ using feature differences relative to a base arm $K$ is necessary because only relative differences in latent utilities drive choice probabilities. This normalization, combined with the full column rank condition, guarantees the global identifiability of the parameters $\theta$ \citep{train2009discrete}.

\begin{remark}[Rank condition]
The full-column-rank condition on $X_{1:T}$ is mild and holds generically
under non-degenerate reward distributions.
In the stationary 2-armed bandit, the covariates
$(V_t,\, RU_t,\, V_t/TU_t)$ vary across rounds because the Kalman-filter
updates in continuously revise
$Q_k(t)$ and $S_k(t)$ as new rewards are observed, so the rank condition
is satisfied almost surely.
In the non-stationary 4-armed bandit, the diffusion process
ensures that the value and uncertainty trajectories are distinct across arms
almost surely, again guaranteeing the rank condition.
\end{remark}

\newpage
\section{Experimental Results}~\label{appendix:results}
We first present the key experimental results supporting our main analysis. Tables \ref{tab:loo_cv_both}, \ref{tab:2armexploitation-rates},~\ref{tab:alg_params_2arm},  ~\ref{tab:alg_params_4arm} and ~\ref{tab:regret_exploit} correspond to key findings referenced in the main text. 

Table~\ref{tab:loo_cv_both} presents individual-level LOO-CV results for both experiments, comparing the predictive performance of different choice models across various LLMs and human data. Since the datasets vary in size, the LOO-CV scores are normalized by the number of data points to allow comparison across conditions. Reported values are the normalized \emph{expected log point-wise predictive density (elpd\_loo)} estimates, as defined in the loo package \citep{vehtari2017practical}. Standard errors are omitted because the scores are normalized per data point and are intended for relative model comparison.

\begin{table}[ht]
\centering
\small
\caption{Normalized Individual-level LOO-CV score comparison. The best model is highlighted in boldface.}
\label{tab:loo_cv_both}
\begin{subtable}[t]{0.4\textwidth}
\centering
\caption{2-armed bandit}
\label{tab:loo_cv_2armed}
\begin{tabular}{lcccc}
\toprule
\textbf{Agent / Prompt} & SM-1 & SM-2 & SM-3 & Probit \\
\midrule
\textsc{ChatGPT-4o-mini} &   -0.385    &    \textbf{-0.381}   &   \textbf{-0.381}    &  -0.382     \\
\textsc{ChatGPT-4o-mini} + CoT &   -0.332    &  \textbf{-0.331}     &  -0.332     &   -0.337    \\
\textsc{ChatGPT-o3-mini} (High-thinking) &    -0.357   &   -0.287    &   -0.229    &   \textbf{-0.178}    \\
\textsc{ChatGPT-o3-mini} (Low-thinking)
&  -0.300   &   -0.278    &   -0.249    &   \textbf{-0.226} \\
\addlinespace
\textsc{Gemini-2.0-flash}  &    -0.334   &   -0.331    &    \textbf{-0.330}   &   -0.338   \\
\textsc{Gemini-2.0-flash} + CoT &    -0.395   &   -0.342    &    -0.334   &   \textbf{-0.311}   \\
\textsc{Gemini-2.5-flash} (High-thinking) &    -0.203   &   -0.199    &    -0.200   &   \textbf{-0.174}    \\
\textsc{Gemini-2.5-flash} (Non-thinking) &    -0.464   &   -0.465    &    -0.461   &   \textbf{-0.449} \\
\addlinespace
\textsc{DeepSeek-V3} &  -0.322     &    \textbf{-0.319}   &   \textbf{-0.319}    &   -0.332    \\
\textsc{DeepSeek-V3} + CoT &  -0.413     &  -0.334     &   -0.324    &   \textbf{-0.314}    \\
\textsc{DeepSeek-R1}  &   -0.360    &    -0.267   &    -0.195   &    \textbf{-0.168}   \\
\addlinespace
Human Agents &   -0.393    &    -0.346   &   -0.335    &   \textbf{-0.329}    \\
UCB & -0.116 & -0.115& -0.114 & \textbf{-0.111}\\
TS &-0.226&	-0.220&	\textbf{-0.220} & -0.224\\
$\epsilon$-greedy &-0.250 &	-0.229	& -0.229 & \textbf{-0.237}\\
\bottomrule
\end{tabular}
\end{subtable}
\hfill
\begin{subtable}[t]{0.4\textwidth}
\centering
\caption{4-armed bandit}
\label{tab:loo_cv_4armed}
\begin{tabular}{lccc}
\toprule
SM-1 & SM-2 & SM-3 \\
\midrule
-1.029    &   -0.059    &   \textbf{-0.058}    \\
  -0.857    &   -0.319    &    \textbf{-0.318}   \\
   -0.567   &   -0.257    &   \textbf{-0.255}    \\
    -0.830   &   -0.184    &   \textbf{-0.180}    \\
\addlinespace
  -0.986    &  -0.107     &    \textbf{-0.105}   \\
  -0.570    &    -0.452   &   \textbf{-0.452}    \\
  -0.528    &    -0.375   &   \textbf{-0.374}    \\
  -0.693    &    -0.154   &   \textbf{-0.153}    \\
\addlinespace
  -0.872   &   \textbf{-0.103}    &   \textbf{-0.103}   \\
    -0.581   &   \textbf{-0.391}    &   \textbf{-0.391}    \\
  -0.484    &   -0.313    &  \textbf{-0.312}     \\
\addlinespace
-0.649    &  -0.633     &   \textbf{-0.614}    \\
 -0.868 & \textbf{-0.181}& \textbf{-0.181}\\
-1.137 &	-0.763&	\textbf{-0.762}\\
-1.00&	-0.506&	\textbf{-0.496}\\
\bottomrule
\end{tabular}
\end{subtable}
\vspace{1em}
\end{table}
Table \ref{tab:exploit_2arm} presents the average exploitation rates across independent trials for each agent type in the 2-armed bandit task. Panel ~\ref{tab:exploit_llms} compares different LLMs, while Panel ~\ref{tab:exploit_baselines} reports the average rates for human participants and algorithmic baselines. We observe that exploitation rates are consistently high across agents and closely match the levels achieved by algorithmic baselines.

\begin{table}[ht]
    \centering
    \small
    \caption{Exploitation rates in the 2-armed bandit task across different agents}
    \label{tab:exploit_2arm}

    \begin{subtable}[t]{0.48\linewidth}
        \centering
        \caption{Different LLM versions}
        \label{tab:exploit_llms}
        \begin{tabular}{lcccc}
            \toprule
            Agent & Basic & CoT & High-Reasoning & Low - Reasoning\\
            \midrule
            \textsc{ChatGPT}  & 0.90 & 0.92 & 0.95 & 0.96\\
            \textsc{Gemini}   & 0.91 & 0.91 & 0.91 & 0.84\\
            \textsc{DeepSeek} & 0.91 & 0.90 & 0.94 & N/A\\
            \bottomrule
        \end{tabular}
    \end{subtable}%
    \begin{subtable}[t]{0.48\linewidth}
        \centering
        \caption{Human and algorithmic baselines}
        \label{tab:exploit_baselines}
        \begin{tabular}{lc}
            \toprule
            Agent          & Exploitation Rate \\
            \midrule
            Human          & 0.89         \\
            UCB            & 0.91\\
            TS      &    0.88      
            \\
            $\epsilon$-greedy &   
            0.91
            \\
            \bottomrule
        \end{tabular}
    \end{subtable}
\end{table}
Table \ref{tab:alg_params_2arm} reports the posterior means and standard deviations for estimated population-level mean parameters of \ref{eq:SM3} model and \ref{eq:probit} model for MAB algorithms in the 2-armed bandit experiment.

Table \ref{tab:alg_params_4arm} reports the posterior means and standard deviations for estimated population-level mean parameters of \ref{eq:SM3} model for MAB algorithms in the 4-armed bandit experiment.

\begin{table}[ht]
\centering
\small
\caption{Estimated population-level mean parameters for MAB algorithms in the 2-armed bandit experiment.}
\label{tab:alg_params_2arm}
\begin{tabular}{@{}l ccc ccc@{}}
\toprule
\textbf{Algorithm} & $\beta$ (SM-3) & $\phi$ (SM-3) & $\rho$ (SM-3) & $w_1$ (Probit) & $w_2$ (Probit) & $w_3$ (Probit) \\
\midrule
UCB&  $0.683~(0.055)$&$1.402~(0.315)$	&$0.604~(0.240)$ &$0.071~(0.042)$&	$0.408~(0.106)$&	$1.056~(0.166)$\\
TS                  &$0.22~ (0.02)$  & $-1.84~( 0.62)$  & $0.80 ~(0.49)$  &$0.06 ~(0.03)$  & $-0.32 ~(0.05)$ & $0.18~(0.10)$ \\
$\epsilon$-greedy & $0.15 ~(0.01)$ & $ -5.63~ (1.04)$ & $0.92~(0.67)$ & $0.10~(0.02)$  & $-0.5~(0.05)$ & $-0.10~(0.05)$ \\
\bottomrule
\end{tabular}
\end{table}

\begin{table}[ht]
\centering
\small
\caption{Estimated population-level mean parameters for MAB algorithms in the 4-armed bandit experiment.}
\label{tab:alg_params_4arm}

\begin{tabular}{@{}l ccc @{}}
\toprule
\textbf{Algorithm} & $\beta$ (SM-3) & $\phi$ (SM-3) & $\rho$ (SM-3) \\
\midrule
UCB & $0.057~(0.006)$&$-9.978~(1.149)$&$	-2.486~(1.807)$\\
TS   & $0.03~(0.01)$ & $-13.17 ~(0.97)$ &   $-4.80~(1.68)$ \\
$\epsilon$-greedy &  $0.13~(0.01)$  & $-0.84~(0.12)$  & $6.02~(0.73)$  \\
\bottomrule
\end{tabular}
\end{table}

As shown in Tables~\ref{tab:alg_params_2arm} and \ref{tab:alg_params_4arm}, benchmark algorithms exhibit distinct parameter patterns that are consistent with their underlying decision rules: In the 2-armed bandit experiment, UCB drives directed exploration through an uncertainty bonus (higher $\phi$, $w_2$ and $w_3$), while TS and $\epsilon-$greedy introduce greater random exploration (lower $\beta$) via prior sampling and stochastic choice mechanisms, respectively. In contrast, reasoning-enhanced LLMs display a mixed strategy, combining elements of both directed and random exploration, which leads to more human-like decision behavior. However, in the 4-armed bandit task, these algorithms also fail to engage in effective exploration.

\subsection{Experiments on Model Misspecification}\label{appendix:mis}
As discussed in Section \ref{sec:mis}, we systematically vary the exploration rate ($\epsilon$) of the $\epsilon$-greedy algorithm. Specifically, under the identical stationary two-armed bandit setting with 300 rounds, we fit the SM-3 model using trajectories generated by the $\epsilon$-greedy algorithm for $\epsilon \in \{0, 0.1, 0.2, \dots, 0.9\}$, and the UCB algorithm for $c = 1,2,4,8$ in ${\sigma}_t(k)\cdot\sqrt{\frac{c\log f(t)}{N_t(k)}}$  
\begin{figure}[ht]
    \centering
    \includegraphics[width=0.75\linewidth]{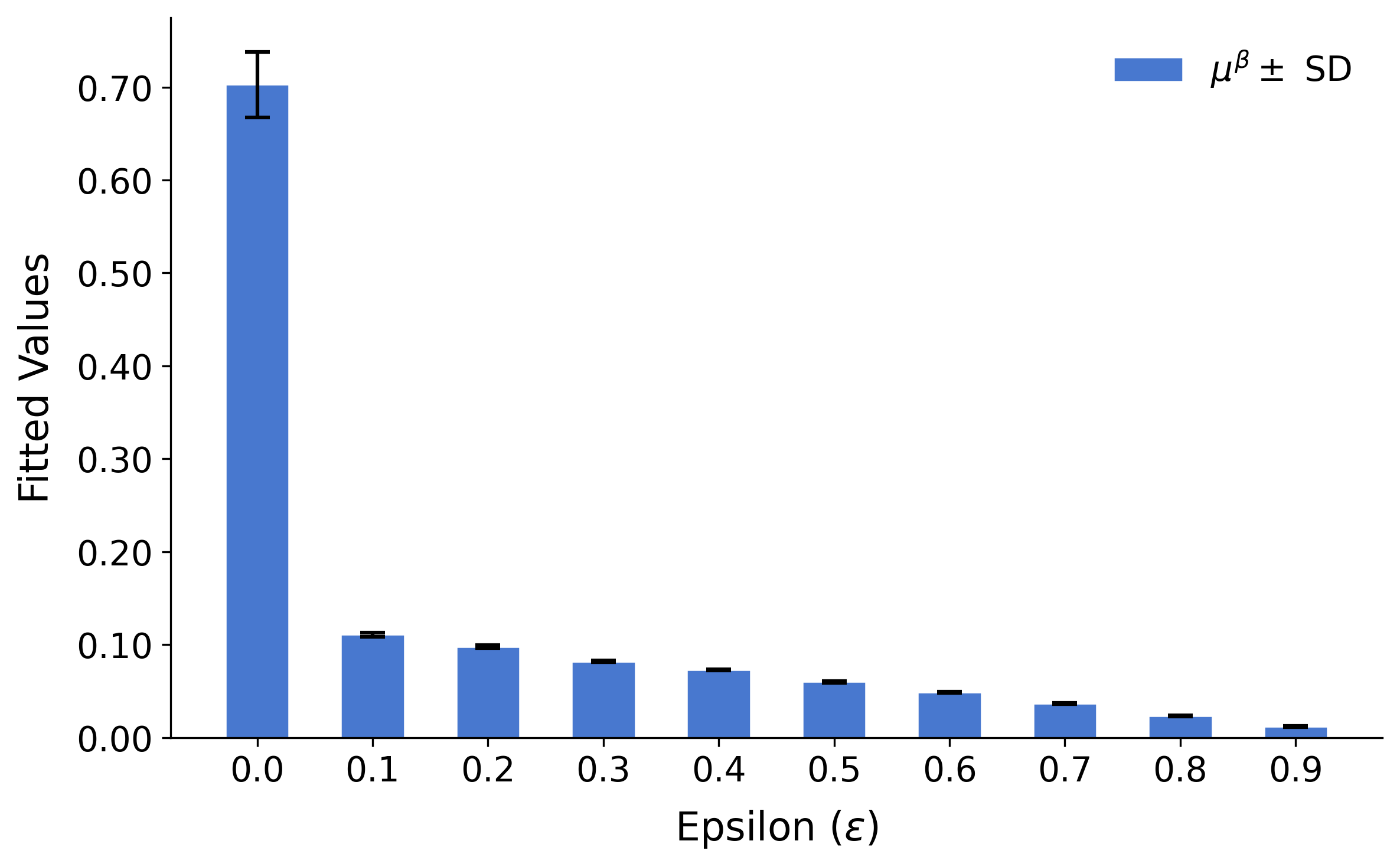}
    \caption{Estimated Random Exploration Parameter ($\beta$) across Varying Exploration Rates ($\epsilon$)}
    \label{fig:change_epsilon}
\end{figure}
\begin{figure}[ht]
    \centering
\includegraphics[width=0.5\linewidth]{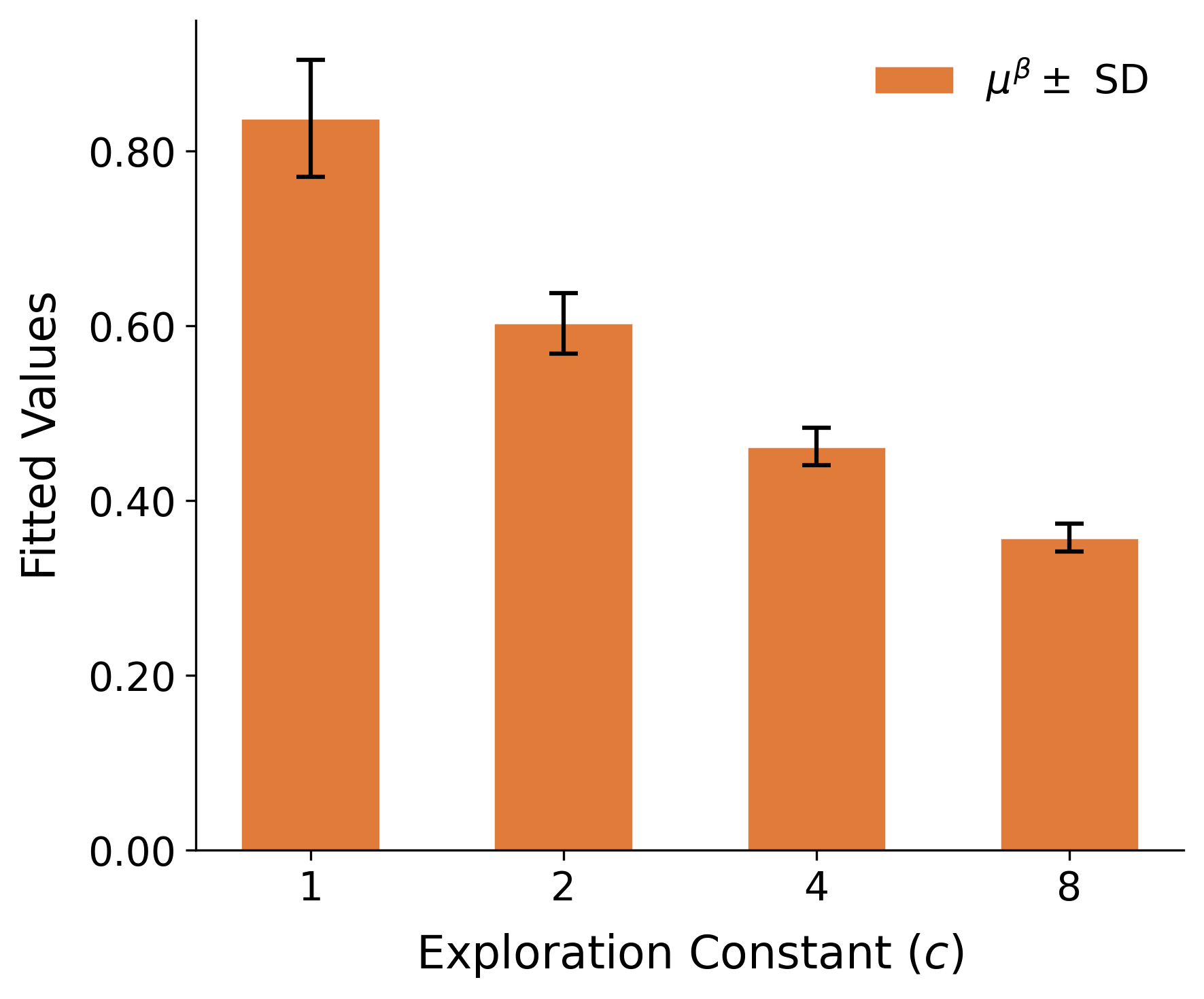}
    \caption{Estimated Random Exploration Parameter across Varying UCB Exploration Constants ($c$)}
    \label{fig:change_c}
\end{figure}
Figure \ref{fig:change_epsilon} illustrates a clear downward trend in the fitted $\beta$ values as $\epsilon$ increases, appropriately reflecting the higher degree of random exploration. Similarly, Figure \ref{fig:change_c} demonstrates that as the UCB exploration $c$ increases, the estimated $\beta$ also exhibits a monotone decrease. Furthermore, Table \ref{tab:eg_params} and \ref{tab:ucb_params} detail the group posterior mean estimates for all three parameters across the varying $\epsilon$ and $c$ values, respectively. 
\begin{table}[ht]
\centering
\small
\caption{Estimated population-level mean parameters for the $\varepsilon$-Greedy algorithm across epsilon values. Values are presented as mean (SD).}
\label{tab:eg_params}
\begin{tabular}{cccc}
\hline
\textbf{$\varepsilon$} & \textbf{$\mu^{\beta}$} & \textbf{$\mu^{\phi}$} & \textbf{$\mu^{\rho}$} \\
\hline
0.0 & 0.703 (0.035) & 3.252 (0.092) & 12.525 (0.487) \\
0.1 & 0.111 (0.002) & -4.530 (0.272) & 9.530 (0.228) \\
0.2 & 0.098 (0.002) & -4.058 (0.372) & 6.876 (0.176) \\
0.3 & 0.082 (0.001) & -4.014 (0.302) & 5.100 (0.153) \\
0.4 & 0.073 (0.001) & -1.840 (0.263) & 3.939 (0.164) \\
0.5 & 0.060 (0.001) & -0.723 (0.299) & 3.006 (0.168) \\
0.6 & 0.049 (0.001) & 2.159 (0.337) & 2.340 (0.168) \\
0.7 & 0.037 (0.001) & 4.572 (0.508) & 1.536 (0.230) \\
0.8 & 0.023 (0.001) & 10.682 (0.888) & 0.765 (0.332) \\
0.9 & 0.012 (0.001) & 27.068 (2.136) & 1.488 (0.621) \\
\hline
\end{tabular}
\end{table}
\begin{table}[ht]
\centering
\small
\caption{Estimated population-level mean parameters for the UCB algorithm across exploration constants $c$ in $\hat{\sigma}_t(k)\cdot\sqrt{\frac{c\log f(t)}{N_t(k)}}$. Values are presented as mean (SD).}
\label{tab:ucb_params}
\begin{tabular}{cccc}
\hline
\textbf{$c$} & \textbf{$\mu^\beta$ (SD)} & \textbf{$\mu^\varphi$ (SD)} & \textbf{$\mu^\rho$ (SD)} \\
\hline
1 & 0.837 (0.067) & 1.780 (0.099) & 4.783 (0.526) \\
2 & 0.603 (0.035) & 1.257 (0.137) & 4.675 (0.524) \\
4 & 0.462 (0.022) & 0.674 (0.134) & 4.684 (0.420) \\
8 & 0.357 (0.016) & 0.927 (0.138) & 6.162 (0.568) \\
\hline
\end{tabular}
\end{table}
For the $\epsilon$-greedy algorithm, the model captures two concurrent behavioral signatures of increasing $\epsilon$: a monotonic decline in the inverse temperature parameter ($\beta$), reflecting greater random exploration, and a parallel decrease in choice perseveration ($\rho$), consistent with more frequent unsystematic arm switching. For the UCB algorithm, the model consistently recovers a positive directed exploration effect ($\phi > 0$) across all values of $c$, while perseveration ($\rho$) remains comparatively stable.
\subsection{Details of fitted values}
In this subsection, we present the details of all fitted values across bandit settings, agents, and choice models.
Tables \ref{tab:4arm_SM3_mu} and \ref{tab:4arm_SM3_sigma} report the posterior means and standard deviations of the population-level parameters estimated from the \ref{eq:SM3} model in the 4-armed bandit task. Table \ref{tab:4arm_SM3_mu} summarizes the population-level means ($\mu^x$) for each parameter $x$, while Table \ref{tab:4arm_SM3_sigma} reports the corresponding population-level standard deviations ($\sigma^x$).
\begin{table}[ht]
\centering
\small
\caption{Posterior means and standard deviations of $\mu^x$ estimated from the SM-3 model in the 4-armed bandit experiment}
\label{tab:4arm_SM3_mu}
\begin{tabular}{lccc}
\toprule
\textbf{Agent/Prompt} & $\mu^{\beta}$ & $\mu^{\phi}$ & $\mu^{\rho}$ \\
\midrule

\textsc{ChatGPT-4o-mini} & 0.094(0.018) & -7.983(1.606) & 6.555(2.819) \\
\textsc{ChatGPT-4o-mini} + CoT & 0.044(0.004) & -10.458(0.995) & 2.404(1.763) \\
\textsc{ChatGPT-o3-mini} (High-thinking) & 0.103(0.005) & -4.190(0.283) & -3.691(0.835) \\
\textsc{ChatGPT-o3-mini} (low-thinking) & 0.073(0.011) & -6.564(0.783) & 12.079(2.364)\\
\addlinespace

\textsc{Gemini-2.0-flash} & 0.082(0.011) & -6.056(0.749) & 10.926(2.827)\\
\textsc{Gemini-2.0-flash} + CoT & 0.096(0.004) & -2.755(0.213) & 0.726(0.670) \\
\textsc{Gemini-2.5-flash} (High-thinking)& 0.109(0.005) & -2.926(0.209) & -2.059(0.632)\\
\textsc{Gemini-2.5-flash} (Non-thinking) & 0.100(0.007) & -4.193(0.374) & 4.640(1.514)\\
\addlinespace
\textsc{DeepSeek-V3} & 0.137(0.016) & -4.419(0.410) & -0.526(1.307) \\
\textsc{DeepSeek-V3} + CoT & 0.092(0.004) & -3.458(0.255) & 0.818(0.780) \\
\textsc{DeepSeek-R1} & 0.135(0.006) & -2.460(0.168) & -2.011(0.546) \\
\addlinespace
Human Agents & 0.168(0.010) & 0.879(0.171) & 5.450(0.299) \\
UCB & 0.057(0.006)&	-9.978(1.149)&	-2.486(1.807)\\
$\epsilon$ - greedy & 0.042(0.003) & -7.774(0.630) & 15.925(1.896) \\
TS & 0.029(0.002) & -13.179(0.972) & -4.797(1.684) \\
\bottomrule
\end{tabular}
\end{table}

\begin{table}[ht]
\centering
\small
\caption{Posterior means and standard deviations of $\sigma^x$ estimated from the SM-3 model in the 4-armed bandit experiment}
\label{tab:4arm_SM3_sigma}
\begin{tabular}{lccc}
\toprule
\textbf{Agent/Prompt} & $\sigma^{\beta}$ & $\sigma^{\phi}$ & $\sigma^{\rho}$ \\
\midrule

\textsc{ChatGPT-4o-mini} & 0.023(0.011) & 0.165(0.115) & 0.184(0.117) \\
\textsc{ChatGPT-4o-mini} + CoT & 0.001(0.001) & 0.183(0.113) & 0.188(0.097) \\
\textsc{ChatGPT-o3-mini} (High-thinking) & 0.004(0.003) & 0.120(0.077) & 0.189(0.114) \\
\textsc{ChatGPT-o3-mini} (Low-thinking) & 0.025(0.013) & 0.221(0.162) & 0.182(0.111)\\
\addlinespace
\textsc{Gemini-2.0-flash} & 0.010(0.005) & 0.173(0.122) & 0.199(0.123)\\
\textsc{Gemini-2.0-flash} + CoT  & 0.002(0.002) & 0.109(0.071) & 0.200(0.111) \\
\textsc{Gemini-2.5-flash} (High-thinking)& 0.003(0.002) & 0.084(0.063) & 0.175(0.109)\\
\textsc{Gemini-2.5-flash} (Non-thinking)& 0.004(0.003) & 0.185(0.132) & 0.202(0.110)\\
\addlinespace
\textsc{DeepSeek-V3} & 0.027(0.016) & 0.312(0.197) & 0.205(0.118) \\
\textsc{DeepSeek-V3} + CoT & 0.003(0.002) & 0.186(0.106) & 0.181(0.111) \\
\textsc{DeepSeek-R1} & 0.006(0.004) & 0.140(0.090) & 0.171(0.114) \\
\addlinespace
Human Agents & 0.053(0.008) & 0.850(0.085) & 0.268(0.162) \\
UCB & 0.006(0.004)&0.182(0.116)&0.188(0.111)\\

$\epsilon$ - greedy & 0.001(0.001) & 0.177(0.116) & 0.217(0.105) \\
TS & 0.003(0.001) & 0.149(0.106) & 0.172(0.107) \\
\bottomrule
\end{tabular}
\end{table}

Tables \ref{tab::4arm_SM2} and \ref{tab::4arm_SM1} present the posterior means and standard deviations of the population-level parameters estimated from the \ref{eq:SM2} and \ref{eq:SM1} models, respectively, in the 4-armed bandit experiment. Table \ref{tab::4arm_SM2} reports the group-level means ($\mu^x$) and standard deviations ($\sigma^x$) for each parameter in the SM-2 model, while Table \ref{tab::4arm_SM1} shows the corresponding estimates for the SM-1 model.
\begin{table}[ht]
\centering
\small
\caption{Posterior means and standard deviations of $\mu^x$ and $\sigma^x$ estimated from the SM-2 model in the 4-armed bandit experiment}
\label{tab::4arm_SM2}
\begin{tabular}{lcccc}
\toprule
\textbf{Agent/Prompt} & $\mu^{\beta}$ & $\mu^{\phi}$ & $\sigma^{\beta}$ & $\sigma^{\phi}$ \\
\midrule
\textsc{ChatGPT-4o-mini} & 0.101(0.022) & -8.862(1.660) & 0.032(0.016) & 0.185(0.115) \\
\textsc{ChatGPT-4o-mini} + CoT & 0.044(0.004) & -10.861(1.081) & 0.002(0.001) & 0.199(0.110) \\
\textsc{ChatGPT-o3-mini} (High-thinking) & 0.098(0.005) & -3.944(0.279) & 0.004(0.002) & 0.123(0.074) \\
\textsc{ChatGPT-o3-mini} (Low-thinking)& 0.068(0.008) & -8.157(0.728) & 0.012(0.006) & 0.357(0.269)\\
\addlinespace
\textsc{Gemini-2.0-flash}& 0.087(0.009) & -6.887(0.617) & 0.010(0.005) & 0.185(0.159) \\
\textsc{Gemini-2.0-flash} + CoT & 0.098(0.004) & -2.761(0.197) & 0.002(0.002) & 0.113(0.072) \\
\textsc{Gemini-2.5-flash} (High-thinking)& 0.105(0.005) & -2.830(0.190) & 0.004(0.002) & 0.104(0.062)\\
\textsc{Gemini-2.5-flash} (Non-thinking) & 0.106(0.007) & -4.467(0.333) & 0.004(0.003) & 0.211(0.125)\\ 
\addlinespace
\textsc{DeepSeek-V3} & 0.138(0.016) & -4.317(0.378) & 0.028(0.017) & 0.327(0.181) \\
\textsc{DeepSeek-V3} + CoT & 0.092(0.004) & -3.540(0.251) & 0.003(0.002) & 0.196(0.108) \\
\textsc{DeepSeek-R1} & 0.133(0.007) & -2.261(0.182) & 0.006(0.004) & 0.121(0.090) \\
\addlinespace
Human Agents & 0.168(0.010) & 0.159(0.153) & 0.051(0.008) & 0.767(0.080) \\
UCB & 0.058(0.006)&-9.303(0.973)&	0.005(0.004)&0.179(0.110)\\
$\epsilon$ - greedy & 0.043(0.003) & -9.975(0.690) & 0.001(0.001) & 0.212(0.121) \\
TS & 0.030(0.002) & -12.325(0.925) & 0.003(0.001) & 0.176(0.103) \\
\bottomrule
\end{tabular}
\end{table}

\begin{table}[ht]
\centering
\small
\caption{Posterior means and standard deviations of $\mu^x$ and $\sigma^x$ estimated from the SM-1 model in the 4-armed bandit task}
\label{tab::4arm_SM1}
\begin{tabular}{lcc}
\toprule
\textbf{Agent/Prompt} & $\mu^{\beta}$ & $\sigma^{\beta}$ \\
\midrule
\textsc{ChatGPT-4o-mini} & 0.142(0.025) & 0.096(0.021) \\
\textsc{ChatGPT-4o-mini} + CoT & 0.150(0.018) & 0.067(0.015) \\
\textsc{ChatGPT-o3-mini} (High-thinking) & 0.243(0.017) & 0.062(0.015) \\
\textsc{ChatGPT-o3-mini} (Low-thinking) & 0.183(0.033) & 0.120(0.025)\\
\addlinespace
\textsc{Gemini-2.0-flash} & 0.163(0.033) & 0.119(0.027) \\
\textsc{Gemini-2.0-flash} + CoT  & 0.187(0.005) & 0.012(0.006) \\
\textsc{Gemini-2.5-flash} (High-thinking)& 0.223(0.009) & 0.027(0.008)\\
\textsc{Gemini-2.5-flash} (Non-thinking) & 0.243(0.032) & 0.122(0.027)\\
\addlinespace
\textsc{DeepSeek-V3} & 0.205(0.038) & 0.140(0.030) \\
\textsc{DeepSeek-V3} + CoT & 0.202(0.013) & 0.048(0.012) \\
\textsc{DeepSeek-R1} & 0.257(0.013) & 0.044(0.011) \\
\addlinespace
Human Agents& 0.161(0.011) & 0.058(0.009) \\
UCB & 0.143(0.013)&0.047(0.011)\\
$\epsilon$ - greedy & 0.106(0.009) & 0.032(0.007) \\
TS & 0.076(0.009) & 0.034(0.008) \\
\bottomrule
\end{tabular}

\end{table}
Tables \ref{tab:2arm_SM3_mu} and \ref{tab:2arm_SM3_sigma} report the posterior means and standard deviations of the population-level parameters estimated from the \ref{eq:SM3} model in the 2-armed bandit experiment. Table \ref{tab:2arm_SM3_mu} summarizes the population-level means ($\mu^x$) for each parameter, while Table \ref{tab:2arm_SM3_sigma} reports the corresponding population-level standard deviations ($\sigma^x$).
\begin{table}[ht]
\centering
\small
\caption{Posterior means and standard deviations of $\mu^x$ estimated from the SM-3 model in the 2-armed bandit experiment}
\label{tab:2arm_SM3_mu}
   \begin{tabular}{lccc}
\toprule
\textbf{Agent/Prompt} & $\mu^{\beta}$ & $\mu^{\phi}$ & $\mu^{\rho}$ \\
\midrule
\textsc{ChatGPT-4o-mini} & 0.206(0.010) & -0.264(0.093) & 0.591(0.402) \\
\textsc{ChatGPT-4o-mini} + CoT & 0.291(0.015) & 0.108(0.071) & -0.099(0.332) \\
\textsc{ChatGPT-o3-mini} (High-thinking) & 0.416(0.025) & 2.562(0.092) & 4.638(0.366) \\
\textsc{ChatGPT-o3-mini} (Low-thinking) & 0.384(0.023) & 1.198(0.079) & 3.905(0.339)\\
\addlinespace
\textsc{Gemini-2.0-flash} & 0.262(0.014) & -0.114(0.074) & 1.104(0.371)\\
\textsc{Gemini-2.0-flash} + CoT & 0.297(0.015) & 1.669(0.109) & 1.936(0.325) \\
\textsc{Gemini-2.5-flash} (High-thinking) &0.449(0.030) & 1.098(0.347) & 0.235(0.301)\\
\textsc{Gemini-2.5-flash} (Non-thinking) & 0.151(0.007) & 0.361(0.110) & 2.409(0.506)\\
\addlinespace
\textsc{DeepSeek-V3} & 0.265(0.016) & -0.220(0.077) & 0.520(0.351) \\
\textsc{DeepSeek-V3} + CoT & 0.319(0.022) & 2.552(0.174) & 2.021(0.322) \\
\textsc{DeepSeek-R1} & 0.590(0.059) & 3.613(0.124) & 4.719(0.300) \\
\addlinespace
Human & 0.436(0.024) & 1.075(0.138) & 1.076(0.273) \\
UCB & 0.683(0.055)&1.402(0.315)	&0.604(0.240)\\
$\epsilon$ - greedy & 0.152(0.013) & -5.631(1.048) & 0.928(0.673) \\
TS & 0.224(0.017) & -1.842(0.625) & 0.796(0.492) \\
\bottomrule
\end{tabular}
\end{table}
\begin{table}[ht]
\centering
\small
\caption{Posterior means and standard deviations of $\sigma^x$ estimated from the SM-3 model in the 2-armed bandit experiment}
\label{tab:2arm_SM3_sigma}
\begin{tabular}{lccc}
\toprule
\textbf{Agent/Prompt} & $\sigma^{\beta}$ & $\sigma^{\phi}$ & $\sigma^{\rho}$ \\
\midrule
\textsc{ChatGPT-4o-mini} & 0.011(0.009) & 0.081(0.060) & 0.328(0.215) \\
\textsc{ChatGPT-4o-mini} + CoT & 0.024(0.014) & 0.090(0.058) & 0.330(0.218) \\
\textsc{ChatGPT-o3-mini} (High-thinking) & 0.025(0.018) & 0.120(0.075) & 0.369(0.262) \\
\textsc{ChatGPT-o3-mini} (Low-thinking)& 0.047(0.026) & 0.229(0.090) & 0.386(0.284)\\
\addlinespace
\textsc{Gemini-2.0-flash} & 0.022(0.014) & 0.089(0.063) & 0.334(0.214)\\
\textsc{Gemini-2.0-flash} + CoT & 0.017(0.012) & 0.257(0.104) & 0.382(0.249) \\
\textsc{Gemini-2.5-flash} (High-thinking) & 0.033(0.023) & 0.365(0.219) & 0.227(0.150)\\
\textsc{Gemini-2.5-flash} (Non-thinking) & 0.008(0.005) & 0.119(0.080) & 0.420(0.300)\\
\addlinespace
\textsc{DeepSeek-V3} & 0.022(0.012) & 0.056(0.046) & 0.347(0.290) \\
\textsc{DeepSeek-V3} + CoT & 0.053(0.024) & 0.498(0.149) & 0.334(0.226) \\
\textsc{DeepSeek-R1} & 0.154(0.065) & 0.120(0.080) & 0.340(0.218) \\
\addlinespace
Human & 0.135(0.022) & 0.849(0.103) & 1.546(0.224) \\
UCB & 0.057(0.047)&0.307(0.206)&0.267(0.178)\\
$\epsilon$ - greedy & 0.019(0.010) & 0.922(0.578) & 0.550(0.421) \\
TS & 0.032(0.015) & 0.433(0.289) & 0.426(0.283) \\
\bottomrule
\end{tabular}
\end{table}

Tables \ref{tab:2arm_SM2} and \ref{tab:2arm_SM1} present the posterior means and standard deviations of the population-level parameters estimated from the \ref{eq:SM2} and \ref{eq:SM1} models, respectively, in the 2-armed bandit experiment. Table \ref{tab:2arm_SM2} reports the group-level means ($\mu^x$) and standard deviations ($\sigma^x$) for each parameter in the SM-2 model, while Table \ref{tab:2arm_SM1} shows the corresponding estimates for the SM-1 model.
\begin{table}[ht]
\centering
\small
\caption{Posterior means and standard deviations of $\mu^x$ and $\sigma^x$ estimated from the SM-2 model in the 2-armed bandit experiment}
\label{tab:2arm_SM2}
\begin{tabular}{lcccc}
\toprule
\textbf{Agent/Prompt} & $\mu^{\beta}$ & $\mu^{\phi}$ & $\sigma^{\beta}$ & $\sigma^{\phi}$ \\
\midrule
\textsc{ChatGPT-4o-mini} & 0.208(0.012) & -0.329(0.072) & 0.012(0.009) & 0.071(0.057) \\
\textsc{ChatGPT-4o-mini} + CoT & 0.291(0.014) & 0.125(0.059) & 0.023(0.014) & 0.090(0.062) \\
\textsc{ChatGPT-o3-mini} (High-thinking) & 0.469(0.024) & 1.555(0.062) & 0.028(0.019) & 0.093(0.066) \\
\textsc{ChatGPT-o3-mini} (Low-thinking) & 0.451(0.032) & 0.660(0.069) & 0.079(0.038) & 0.178(0.068)\\
\addlinespace
\textsc{Gemini-2.0-flash} & 0.275(0.016) & -0.255(0.062) & 0.030(0.014) & 0.082(0.065)\\
\textsc{Gemini-2.0-flash} + CoT & 0.324(0.015) & 1.344(0.089) & 0.019(0.013) & 0.227(0.100) \\
\textsc{Gemini-2.5-flash} (High-thinking) & 0.449(0.030) & 0.940(0.281) & 0.037(0.023) & 0.341(0.250)\\
\textsc{Gemini-2.5-flash} (Non-thinking) & 0.161(0.007) & 0.053(0.079) & 0.008(0.006) & 0.106(0.078)\\
\addlinespace
\textsc{DeepSeek-V3} & 0.274(0.014) & -0.266(0.065) & 0.024(0.016) & 0.065(0.046) \\
\textsc{DeepSeek-V3} + CoT & 0.347(0.025) & 2.098(0.126) & 0.067(0.028) & 0.361(0.122) \\
\textsc{DeepSeek-R1} & 0.532(0.034) & 2.006(0.072) & 0.063(0.038) & 0.103(0.068) \\
\addlinespace
Human & 0.463(0.027) & 0.873(0.129) & 0.151(0.024) & 0.828(0.104) \\
UCB & 0.674(0.056)&	0.888(0.254)&0.075(0.050)&0.333(0.207)\\

$\epsilon$ - greedy & 0.152(0.013) & -6.305(0.982) & 0.020(0.010) & 0.850(0.592) \\
TS & 0.222(0.017) & -2.448(0.518) & 0.032(0.015) & 0.421(0.295) \\
\bottomrule
\end{tabular}
\end{table}
\begin{table}[ht]
\centering
\small
\caption{Posterior means and standard deviations of $\mu^x$ and $\sigma^x$ estimated from the SM-1 model in the 2-armed bandit experiment}
\label{tab:2arm_SM1}
\begin{tabular}{lcc}
\toprule
\textbf{Agent/Prompt} & $\mu^{\beta}$ & $\sigma^{\beta}$ \\
\midrule
\textsc{ChatGPT-4o-mini} & 0.232(0.010) & 0.016(0.011) \\
\textsc{ChatGPT-4o-mini} + CoT & 0.277(0.011) & 0.018(0.012) \\
\textsc{ChatGPT-o3-mini} (High-thinking) & 0.247(0.010) & 0.011(0.008) \\
\textsc{ChatGPT-o3-mini} (Low-thinking) & 0.309(0.016) & 0.030(0.018)\\
\addlinespace
\textsc{Gemini-2.0-flash}& 0.290(0.015) & 0.033(0.016)\\
\textsc{Gemini-2.0-flash} + CoT & 0.207(0.008) & 0.009(0.007) \\
\textsc{Gemini-2.5-flash} (High-thinking) & 0.386(0.020) & 0.031(0.021)\\
\textsc{Gemini-2.5-flash} (Non-thinking) & 0.161(0.007) & 0.010(0.007)
\\
\addlinespace
\textsc{DeepSeek-V3} & 0.300(0.015) & 0.031(0.016) \\
\textsc{DeepSeek-V3} + CoT & 0.190(0.008) & 0.010(0.008) \\
\textsc{DeepSeek-R1} & 0.241(0.010) & 0.012(0.008) \\
\addlinespace
Human & 0.313(0.014) & 0.080(0.013) \\
UCB & 0.556(0.032)&0.038(0.027)\\
$\epsilon$ - greedy & 0.260(0.024) & 0.072(0.026) \\
TS & 0.286(0.021) & 0.061(0.022) \\
\bottomrule
\end{tabular}
\end{table}

Tables \ref{tab:2arm_probit_mu} and \ref{tab:2arm_probit_sigma} report the posterior means and standard deviations of the population-level parameters estimated from the \ref{eq:probit} model in the 2-armed bandit task. Table \ref{tab:2arm_probit_mu} summarizes the population-level means ($\mu^x$) for each parameter, while Table \ref{tab:2arm_probit_sigma} reports the corresponding population-level standard deviations ($\sigma^x$).
\begin{table}[ht]
\centering
\small
\caption{Posterior means and standard deviations of $\mu^x$ estimated from the probit model in the 2-armed bandit experiment}
\label{tab:2arm_probit_mu}
\begin{tabular}{lccc}
\toprule
\textbf{Agent/Prompt} & $\mu^{w_1}$ & $\mu^{w_2}$ & $\mu^{w_3}$ \\
\midrule
\textsc{ChatGPT-4o-mini} & 0.162(0.012) & -0.028(0.010) & -0.181(0.040) \\
\textsc{ChatGPT-4o-mini} + CoT & 0.195(0.014) & 0.033(0.012) & -0.169(0.048) \\
\textsc{ChatGPT-o3-mini} (High-thinking) & 0.003(0.002) & 0.383(0.024) & 1.860(0.092) \\
\textsc{ChatGPT-o3-mini} (Low-thinking) & -0.106(0.022) & 0.073(0.013) & 2.336(0.195)\\
\addlinespace
\textsc{Gemini-2.0-flash} & 0.135(0.014) & -0.042(0.010) & 0.046(0.052)\\
\textsc{Gemini-2.0-flash} + CoT & 0.016(0.013) & 0.146(0.021) & 0.720(0.050) \\
\textsc{Gemini-2.5-flash} (High-thinking) & -0.363(0.059) & 0.250(0.111) & 2.114(0.230)\\
\textsc{Gemini-2.5-flash} (Non-thinking)& 0.019(0.007) & -0.012(0.007) & 0.317(0.033)\\
\addlinespace
\textsc{DeepSeek-V3} & 0.355(0.027) & -0.007(0.014) & -0.711(0.078) \\
\textsc{DeepSeek-V3} + CoT & 0.001(0.015) & 0.205(0.023) & 0.638(0.057) \\
\textsc{DeepSeek-R1} & 0.006(0.004) & 0.660(0.066) & 1.952(0.217) \\
\addlinespace
Human & 0.025(0.021) & 0.140(0.023) & 0.877(0.091) \\
UCB & 0.071(0.042)&	0.408(0.106)&	1.056(0.166)\\
$\epsilon$ - greedy & 0.101(0.018) & -0.563(0.051) & -0.096(0.057) \\
TS & 0.064(0.029) & -0.321(0.053) & 0.178(0.099) \\
\bottomrule
\end{tabular}
\end{table}

\begin{table}[ht]
\centering
\small
\caption{Posterior means and standard deviations of $\sigma^x$ estimated from the probit model in the 2-armed bandit experiment}
\label{tab:2arm_probit_sigma}
\begin{tabular}{lccc}
\toprule
Agent/Prompt & $\sigma^{w_1}$ & $\sigma^{w_2}$ & $\sigma^{w_3}$ \\
\midrule
\textsc{ChatGPT-4o-mini} & 0.010(0.006) & 0.011(0.008) & 0.028(0.019) \\
\textsc{ChatGPT-4o-mini} + CoT & 0.013(0.008) & 0.018(0.012) & 0.037(0.025) \\
\textsc{ChatGPT-o3-mini} (High-thinking) & 0.001(0.001) & 0.021(0.014) & 0.067(0.050) \\
\textsc{ChatGPT-o3-mini} (Low-thinking) & 0.013(0.011) & 0.023(0.014) & 0.245(0.112)\\
\addlinespace
\textsc{Gemini-2.0-flash} & 0.023(0.010) & 0.015(0.010) & 0.059(0.038)\\
\textsc{Gemini-2.0-flash} + CoT & 0.012(0.007) & 0.019(0.015) & 0.070(0.044) \\
\textsc{Gemini-2.5-flash} (High-thinking)& 0.015(0.011) & 0.298(0.095) & 0.063(0.047)\\
\textsc{Gemini-2.5-flash} (Non-thinking) & 0.006(0.004) & 0.009(0.006) & 0.034(0.021)\\
\addlinespace
\textsc{DeepSeek-V3} & 0.019(0.012) & 0.020(0.014) & 0.033(0.028) \\
\textsc{DeepSeek-V3} + CoT & 0.030(0.012) & 0.057(0.024) & 0.102(0.049) \\
\textsc{DeepSeek-R1} & 0.002(0.002) & 0.166(0.066) & 0.671(0.164) \\
\addlinespace
Human & 0.113(0.019) & 0.139(0.020) & 0.498(0.080) \\
UCB & 0.019(0.015)&	0.133(0.087)& 0.133(0.081)
 \\
$\epsilon$ - greedy & 0.012(0.007) & 0.096(0.056) & 0.028(0.026) \\
TS & 0.022(0.012) & 0.057(0.040) & 0.069(0.039) \\
\bottomrule
\end{tabular}
\end{table}
Table~\ref{tab:regret_exploit} reports the average cumulative regret and exploitation rates in the final round of both bandit experiments. Results are presented for a range of LLM agents under various prompting conditions, including human participants and the UCB algorithm. Standard errors are provided for regret. The comparison highlights systematic differences in performance across models, thinking modes, and task complexity.
\begin{table}[ht]
\centering
\small
\caption{Average cumulative regret and exploitation rates in the last round across both experiments, with standard errors reported for regret.}
\label{tab:regret_exploit}
\begin{tabular}{lcccc}
\toprule
\multicolumn{1}{c}{} & \multicolumn{2}{c}{\textbf{2-armed bandit task}} & \multicolumn{2}{c}{\textbf{4-armed bandit task}} \\
\textbf{Agent/Prompt} & Regret & Exploit Rate & Regret & Exploit Rate \\
\midrule

\textsc{ChatGPT-4o-mini} & 13.467(0.911) & 0.897 & 3599.267(352.296) & 0.653 \\
\textsc{ChatGPT-4o-mini}+CoT & 12.65(0.725) & 0.921 & 2472.733(320.921) & 0.698 \\

\textsc{ChatGPT-o3-mini} (High-thinking)& 14.143(0.492) & 0.952 & 1803.6(235.734) & 0.821 \\

\textsc{ChatGPT-o3-mini} (Low-thinking) & 11.923(0.528) & 0.962 & 3064.133(432.816) & 0.708 \\

\addlinespace

\textsc{Gemini-2.0-flash} & 10.963(0.613) & 0.905 & 3685.8(292.369) & 0.66 \\
\textsc{Gemini-2.0-flash}+ CoT & 15.653(0.448) & 0.914 & 1504.4(169.034) & 0.814 \\
\textsc{Gemini-2.5-flash} (High-thinking) & 15.74(0.437) & 0.912 & 1551.067(171.668) & 0.84 \\
\textsc{Gemini-2.5-flash} (Low-thinking) & 19.713(0.738) & 0.84 & 2551.0(309.847) & 0.762 \\
\addlinespace

\textsc{DeepSeek-V3} & 10.943(0.693) & 0.908 & 3089.467(314.202) & 0.7 \\
\textsc{DeepSeek-V3} + CoT & 16.55(0.556) & 0.903 & 1678.133(216.983) & 0.802 \\
\textsc{DeepSeek-R1}  & 13.987(0.464) & 0.943 & 1532.4(246.638) & 0.838 \\
\addlinespace
\textsc{Human} & 11.281(0.398) & 0.893 & 1509.355(125.098) & 0.768 \\
\textsc{UCB} & 14.723(0.446) & 0.908 & 2307.933(134.985) & 0.698 \\
\bottomrule
\end{tabular}
\end{table}

\section{LLM versions, dates, and API endpoints}\label{appendix:versions}
Tables \ref{tab:model_versions_endpoints_main_experements}, \ref{tab:model_versions_2026}, and \ref{tab:model_versions_2arms_300_rounds} show the model versions, dates, and API for each model used in the experiments. Most models were invoked through the \texttt{openai} Python library (version~1.66.2). In the 2-armed bandit experiment with 300 rounds, the \textsc{Gemini-2.5-flash} model was invoked through \texttt{google-genai} library (version~3.0.0), and the \textsc{Gemini-2.0-flash} model was invoked by directly posting to OpenRouter's API endpoint, with the model specified as \texttt{google/gemini-2.0-flash-001} in the request body.
\begin{table}[ht]
\centering
\small
\begin{tabular}{lllp{0.36\linewidth}}
\hline
\textbf{Model name} & \textbf{Date} & \textbf{Version} & \textbf{API endpoint} \\
\hline
\texttt{gpt-4o-mini} 
& Mar. 20, 2025 
& \texttt{gpt-4o-mini-2024-07-18} 
& \url{https://api.openai.com/v1/chat/completions} \\

\texttt{o3-mini} 
& Apr. 2, 2025 
& \texttt{o3-mini-2025-01-31} 
& \url{https://api.openai.com/v1/chat/completions} \\

\texttt{gemini-2.5-flash} 
& Aug. 4, 2025 
& \texttt{gemini-2.5-flash} 
& \url{https://generativelanguage.googleapis.com/v1beta/openai/chat/completions} \\

\texttt{gemini-2.0-flash} 
& Apr. 2, 2025 
& \texttt{gemini-2.0-flash-001} 
& \url{https://generativelanguage.googleapis.com/v1beta/openai/chat/completions} \\

\texttt{deepseek-reasoner} 
& Mar. 22, 2025 
& \texttt{DeepSeek-R1} 
& \url{https://api.deepseek.com/chat/completions} \\

\texttt{deepseek-chat} 
& Apr. 10, 2025 
& \texttt{DeepSeek-V3-0324} 
& \url{https://api.deepseek.com/chat/completions} \\
\hline
\end{tabular}
\caption{Model names, experiment dates, model versions, and API endpoints used in the main experiments.}
\label{tab:model_versions_endpoints_main_experements}
\end{table}

\begin{table}[ht]
\centering
\small
\begin{tabular}{lllp{0.36\linewidth}}
\hline
\textbf{Model name} & \textbf{Experiment date} & \textbf{Version} & \textbf{API endpoint} \\
\hline

\texttt{gpt-4o-mini}
& Mar 31, 2026
& \texttt{gpt-4o-mini-2024-07-18}
& \url{https://api.openai.com/v1/chat/completions} \\

\texttt{gemini-2.0-flash}
& Mar 17, 2026
& \texttt{gemini-2.0-flash-001}
& \url{https://openrouter.ai/api/v1/chat/completions} \\

\texttt{deepseek-chat}
& Mar 17, 2026
& \texttt{DeepSeek-V3.2}
& \url{https://api.deepseek.com/chat/completions} \\

\hline
\end{tabular}
\caption{Model names, experiment dates, model versions, and API endpoints used in the experiments with high and low temperatures.}
\label{tab:model_versions_2026}
\end{table}
\begin{table}[ht]
\centering
\small
\begin{tabular}{lllp{0.36\linewidth}}
\hline
\textbf{Model name} & \textbf{Experiment date} & \textbf{Version} & \textbf{API endpoint} \\
\hline

\texttt{o3-mini}
& Feb 26, 2026
& \texttt{o3-mini-2025-01-31}
& \url{https://api.openai.com/v1/chat/completions} \\

\texttt{gpt-4o}
& Feb 26, 2026
& \texttt{gpt-4o-mini-2024-07-18}
& \url{https://api.openai.com/v1/chat/completions} \\

\texttt{deepseek-reasoner}
& Jan 6, 2026
& \texttt{DeepSeek-V3.2}
& \url{https://api.deepseek.com/chat/completions} \\

\texttt{deepseek-chat}
& Feb 24, 2026
& \texttt{DeepSeek-V3.2}
& \url{https://api.deepseek.com/chat/completions} \\

\texttt{gemini-2.0-flash}
& Mar 4, 2026
& \texttt{gemini-2.0-flash-001}
& \url{https://openrouter.ai/api/v1/chat/completions} \\

\texttt{gemini-2.5-flash}
& Mar 4, 2026
& \texttt{gemini-2.5-flash}
& \url{https://generativelanguage.googleapis.com/v1beta/models/gemini-2.5-flash:generateContent} \\

\hline
\end{tabular}
\caption{Model names, experiment dates, model versions, and API endpoints used in the 2-armed bandit experiments with 300 rounds.}
\label{tab:model_versions_2arms_300_rounds}
\end{table}
\end{APPENDICES}
\end{document}